\pdfoutput=1

\documentclass[10pt,twocolumn,letterpaper,dvipsnames,table]{article}

\usepackage{cvpr}              
\usepackage{makecell}
\usepackage{amssymb} 
\usepackage[accsupp]{axessibility}  
\usepackage{graphicx}
\usepackage{algorithm}
\usepackage{algpseudocode}
\usepackage{amsmath}
\usepackage{amssymb}
\usepackage{marvosym}
\usepackage{footmisc}
\usepackage{xspace}
\usepackage{multirow}

\newcommand{\ours}{AdaMMS\xspace}

\makeatletter
\g@addto@macro\normalsize{%
  \setlength{\abovedisplayskip}{5pt}%
  \setlength{\belowdisplayskip}{5pt}%
  \setlength{\abovedisplayshortskip}{3pt}%
  \setlength{\belowdisplayshortskip}{3pt}%
}
\makeatother

\definecolor{cvprblue}{rgb}{0.21,0.49,0.74}

\usepackage{hyperref}
\hypersetup{pagebackref,breaklinks,colorlinks,allcolors=cvprblue}

\def\paperID{8339} 
\def\confName{CVPR}
\def\confYear{2025}

\title{AdaMMS: Model Merging  for Heterogeneous Multimodal Large Language Models with Unsupervised Coefficient Optimization }

\author{
Yiyang Du\textsuperscript{*,1}, 
Xiaochen Wang\textsuperscript{*,3,4}, 
Chi Chen\textsuperscript{*,1}, 
Jiabo Ye\textsuperscript{5},
Yiru Wang\textsuperscript{8}, \\ 
{
Peng Li \textsuperscript{\Letter,2,6}, 
Ming Yan\textsuperscript{5}, 
Ji Zhang\textsuperscript{5}, 
Fei Huang\textsuperscript{5}, 
Zhifang Sui\textsuperscript{3}, 
Maosong Sun\textsuperscript{1}, 
Yang Liu\textsuperscript{\Letter,1,2,6,7}} \\
  \textsuperscript{1}Dept. of Comp. Sci. \& Tech., Institute for AI, Tsinghua University, Beijing, China \\
  \textsuperscript{2}Institute for AI Industry Research (AIR), Tsinghua University, Beijing, China \\
  \textsuperscript{3}State Key Laboratory of Multimedia Information Processing, Peking University, Beijing, China\\
  \textsuperscript{4}School of Software Microelectronics, Peking University, Beijing, China\\
  \textsuperscript{5}Institute of Intelligent Computing, Alibaba Group\\
  \textsuperscript{6}Shanghai Artificial Intelligence Laboratory, Shanghai, China\\
  \textsuperscript{7}Jiangsu Collaborative Innovation Center for Language Competence, Jiangsu, China\\
  \textsuperscript{8}ModelTC Open Source Organization, Beijing, China
}

\begin{document}
\maketitle
\begin{abstract}
\renewcommand{\thefootnote}{\fnsymbol{footnote}} 
    \footnotetext[1]{Equal contribution.}
\DefineFNsymbols*{1}{\Letter}
\setfnsymbol{1}

\renewcommand{\thefootnote}{\fnsymbol{footnote}} 
    \footnotetext[1]{Corresponding authors: Peng Li (lipeng@air.tsinghua.edu.cn) and Yang Liu (liuyang2011@tsinghua.edu.cn).}
\renewcommand{\thefootnote}{\arabic{footnote}}

Recently, model merging methods have demonstrated powerful strengths in combining abilities on various tasks from multiple Large Language Models (LLMs). While previous model merging methods mainly focus on merging homogeneous models with identical architecture, they meet challenges when dealing with Multimodal Large Language Models (MLLMs) with inherent heterogeneous property, including differences in model architecture and the asymmetry in the parameter space. In this work, we propose \ours\footnote{AdaMMS represents \textbf{Ada}ptive \textbf{M}apping, \textbf{M}erging, and \textbf{S}earching.}, a novel model merging method tailored for heterogeneous MLLMs. Our method tackles the challenges in three steps: mapping, merging and searching. Specifically, we first design mapping function between models to apply model merging on MLLMs with different architecture. Then we apply linear interpolation on model weights to actively adapt the asymmetry in the heterogeneous MLLMs. Finally in the hyper-parameter searching step, we propose an unsupervised hyper-parameter selection method for model merging. As the first model merging method capable of merging heterogeneous MLLMs without labeled data, extensive experiments on various model combinations demonstrated that \ours outperforms previous model merging methods on various vision-language benchmarks.\footnote{Code at \url{https://github.com/THUNLP-MT/AdaMMS}.}


\end{abstract}    
\section{Introduction}
\label{sec:intro}



\begin{figure*}
    \centering
    \includegraphics[width=0.9\linewidth, bb=0 0 1074 498]{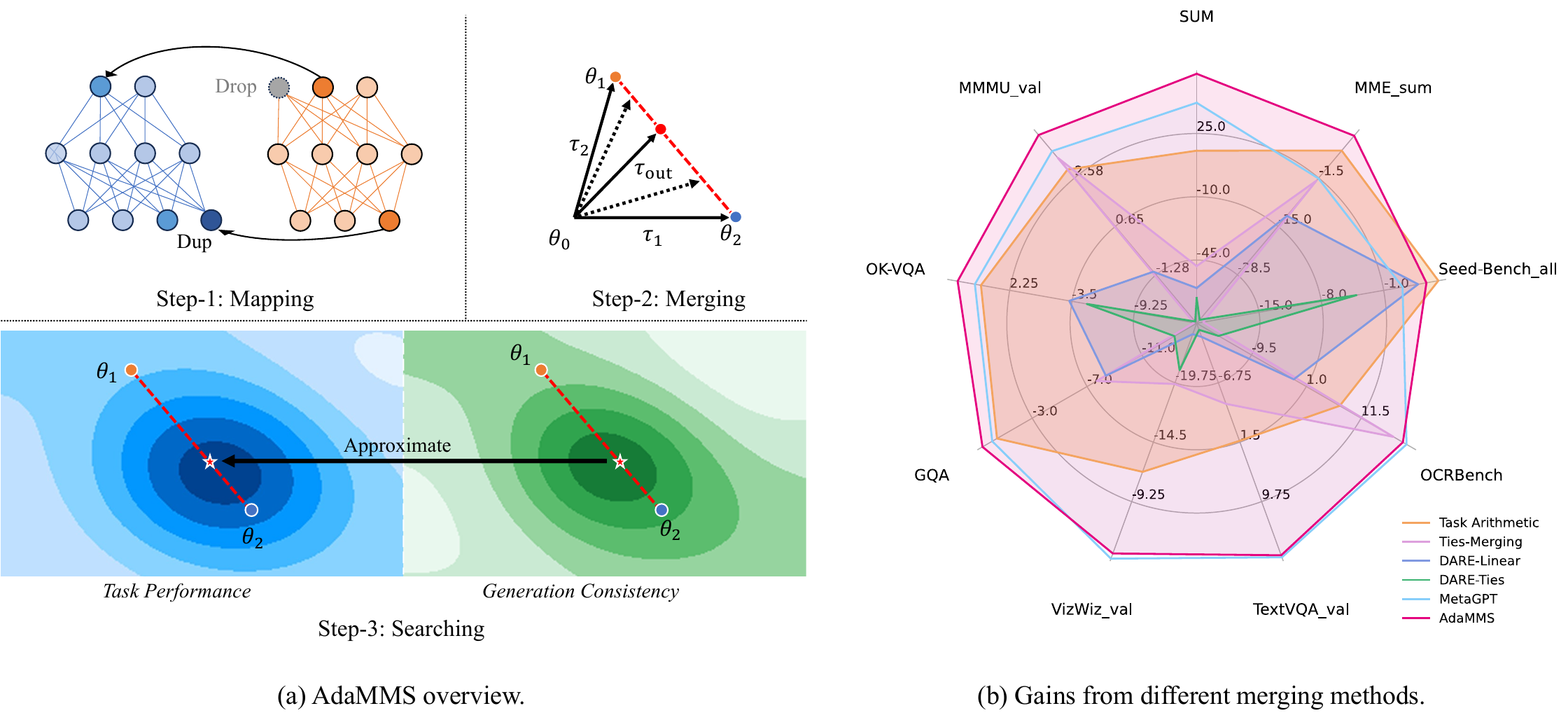}
    \caption{(a) Illustration of three steps in AdaMMS: Step-1, mapping MLLMs with different model architecture; Step-2, merging MLLMs with linear interpolation; Step-3, searching for optimal merging hyper-parameter by approximate task performance through generation consistency without labeled data. (b) The gain performance of AdaMMS on a broad range of multimodal tasks in comparison with existing merging approaches. Gain refers to the improvement obtained by subtracting the average result from the result of the fused model on a certain task. The result here is the average of the gains from the two MLLM pairs merging. }
    \label{fig:figure1}
\end{figure*}

   
        
       


Model merging~\cite{task-arithmetic} has gained increasing popularity in the field of large language models~(LLMs)~\cite{task-arithmetic, ties, dare, metagpt}. This approach typically involves combining the parameters of two models with the same architecture, creating a new model without requiring additional training~\cite{task-arithmetic}. It has proven to be an efficient method for developing models that integrate the abilities of multiple existing models~\cite{ties, dare, metagpt}, avoiding the need for extensive data and computational resources, and has been widely adopted in building powerful LLMs~\cite{goddard2024mergekit,open-llm-leaderboard-v2}.


Despite its popularity with LLMs, model merging has yet to be widely adopted for multimodal large language models~(MLLMs). Some recent studies have explored the application of model merging to MLLMs~\cite{sung2023empirical,modelcompose} but either prioritize extending multimodal capabilities over enhancing the performance of existing models~\cite{modelcompose} or still require additional training after merging~\cite{sung2023empirical}. The primary obstacle in applying model merging to MLLMs lies in the heterogeneous nature of these models~\cite{llava1.5, qwen2-vl, cogvlm2, llava-onevison, mplugowl2}.  This heterogeneity arises from modifications to the transformer architectures~\cite{attention} within their language models~\cite{cogvlm2,mplugowl2}, as well as differences in their choice of modality-specific encoders and tokenizers~\cite{qwen2-vl, cogvlm2}. Consequently, a significant challenge in merging MLLMs is that the process cannot be directly applied to models with different architectures, as their weights are not isomorphic.

Recent efforts like FuseLLM~\cite{fusellm} and FuseChat~\cite{fusechat} have explored fusing the capabilities of heterogeneous LLMs by merging their generative distributions. Theoretically, these methods could also be applied to MLLMs. However, they rely on supervised continued training, which incurs significant computational costs and fails to address scenarios where labeled data is unavailable. For example, FuseLLM requires a training dataset with a total of 1.8 billion tokens and 33 hours of training time. This underscores the need for an unsupervised model merging technique to effectively integrate heterogeneous MLLMs.

In this work, we address the challenges of merging heterogeneous MLLMs by introducing a novel model merging strategy named \ours
, as illustrated in Figure~\ref{fig:figure1}.
First, to enable model merging across heterogeneous MLLMs with differing architectures, we design a parameter mapping framework. Specifically, we focus on the scenario where MLLMs have different language model architectures due to variations in transformer block duplications \cite{cogvlm, cogvlm2, mplugowl2}. This mapping aligns corresponding components across models, enabling merging operations even with structural differences. Next, we apply adaptive linear interpolation on the mapped parameters during merging. By adjusting the interpolation coefficient, \ours optimizes performance adaptively across different tasks. This coefficient adjustment is then optimized through an unsupervised procedure in the following step. Finally, we introduce an \textbf{unsupervised} hyperparameter selection method to determine the interpolation coefficient for each task. This approach leverages response consistency across candidate models as a performance estimate, based on our novel insight that model performance correlates with generation consistency. In addition to strong empirical support, we provide a theoretical analysis to explain this insight.

Extensive experiments have demonstrated the effectiveness of our proposed \ours in merging MLLMs. Our main experiments are conducted on two pairs of heterogeneous MLLMs, one of them is based on Qwen \cite{qwen2} architecture, another pair is based on LLaMA \cite{llama} architecture. Both experiments shows that our model can effectively combine the capabilities of heterogeneous MLLMs by enabling model merging on models with different architecture. The experiments also demonstrated that the merging strategy of \ours, accompanied with our unsupervised hyper-parameter selection method, outperforms previous model merging methods on various vision-language tasks. We also conducted experiments to show that the unsupervised hyper-parameter selection method can be performed with fewer unlabeled data without harming its performance, which shows the robustness of the method and further decrease the data requirements of \ours.


Our main contributions are as follows:

\begin{itemize}

    \item We introduce a novel model-merging strategy, \ours, designed to address the challenges of merging heterogeneous MLLMs. By defining a parameter mapping between different models, \ours facilitates model merging techniques even when architectural differences exist.

    \item We propose an \textit{unsupervised} hyperparameter selection method inspired by the observation that model performance can be effectively estimated through generation consistency—measured by the variation in generated responses. Unlike previous approaches requiring labeled data, this method eliminates the dependency on annotations and can be applied to a small subset of 100 target sample without sacrificing effectiveness.

    \item Comprehensive experiments conducted on various model pairs demonstrate the effectiveness of our approach. Specifically, evaluations on Qwen-based and LLaVA-based heterogeneous MLLM pairs show that \ours successfully combines capabilities from distinct MLLMs. Across various vision-language tasks, \ours consistently outperforms baseline strategies, achieving superior overall performance.

\end{itemize}

\section{Related Work}
\label{sec:bk}

\subsection{Model Merging}

Recent researches on model merging techniques have contributed to building more capable models by providing an efficient approach to combine abilities on various tasks from different models that requires less data and compute.
Some studies focus on merging homogeneous models with identical architecture, while others focus on tackle the challenge on heterogeneous models which have different architectures.
In merging homogeneous models, Task Arithmetic \cite{task-arithmetic} proposes the concept of task vectors, which subtracts fine-tuned weights from pre-train weights to obtain task-related weight difference as the object of merging. Ties-Merging \cite{ties} and DARE \cite{dare} further improve the performance by mitigating parameter interference during the merging process through parameter pruning and conflict resolving. MetaGPT \cite{metagpt} scales the task vectors with task-agnostic coefficients in closed-form by seperating data term and scaling coefficients in the optimization objective. Although these methods improves the performance of the merged models, they cannot be directly applied on models with architecture difference.
In fusing heterogeneous models, DAMC \cite{modelcompose} employs parameter decoupling and adaptive adjustment to enhance model merging strategies for fusing modalities on MLLMs with different modality encoders, but this work still focus on merging identical language model architecture. To consolidate LLMs with different architectures, FuseLLM \cite{fusellm} and FuseChat \cite{fusechat} applies token alignment and model fusion strategies with knowledge distillation before continue training the model, but they need labeled data and computation resources for continue training.
In fact, the majority of previous works on model merging requires labeled data for validation search or supervised training \cite{task-arithmetic, ties, dare, modelcompose, fusellm, fusechat}. In this work, we eliminate the need of labeled data by leveraging our unsupervised hyper-parameter selection method, and enable model merging strategies to be applied on heterogeneous MLLMs with architecture differences.


\subsection{Multimodal Large Language Models} \label{mllms}

As large language models demonstrate huge success in obtaining great abilities in general, recent researches on MLLMs have successfully appending multimodal processing and generation ability on LLMs, especially on the vision modality \cite{llava1.5, sharegpt4v, mplugowl2, cogvlm, qwen2-vl, llava-onevison}. However, these models often adapts unique modifications on language model architecture, resulting in a set of heterogeneous MLLMs, which prevent model merging methods to be applied on them. Specifically, there are two levels of architecture differences among MLLMs. 
First, two MLLMs may be designed from different pre-trained language model. For example, Qwen2-VL \cite{qwen2-vl} and LLaVA-OneVision-Qwen \cite{llava-onevison} are designed from Qwen2 \cite{qwen2}, while LLaVA \cite{llava1.5}, mPLUG-Owl2 \cite{mplugowl2}, CogVLM \cite{cogvlm} and ShareGPT4V \cite{sharegpt4v} are designed following the LLaMA \cite{llama} architecture. 
Second, two MLLMs developed from the same pre-trained language model can still be heterogeneous because they are designed with different modifications on the language model. For example, although CogVLM and mPLUG-Owl2 are both developed from LLaMA architecture, CogVLM adapts visual experts by duplicating query, key and value weights in attention head, while mPLUG-Owl2 is designed to duplicate key, value, and layer norm weights instead. 
The first level of differences is hard to merge, since model merging applies to parameters that are trained from the same pre-training weights \cite{task-arithmetic}. In this work, we tackle the second level of architecture differences via our proposed \ours method.

\section{Method}
\label{sec:method}
To tackle the heterogeneous challenges in merging MLLMs, we propose a novel model merging method named \ours. As shown in Figure~\ref{fig:figure1}(a), it involves three steps: mapping, merging and searching. In the mapping step, we define a mapping function that enables the merging of parameters from different architectures. Next, in the merging step, we apply linear interpolation to adaptively optimize the performance on specific downstream tasks. Finally, in the searching step, we design a unsupervised hyper-parameter selection method for choosing linear interpolation coefficient during merging. This method is based on our novel discovery that the model performance in the parameter space can be approximated by the difference among model responses without the need of labeled data.
\subsection{Mapping} \label{mapping}


To merge the parameters of two heterogeneous models $M_1$ and $M_2$ into $M_1$'s architecture, we need to align their parameters by defining a mapping $f$ that maps each parameter $\theta_1$ in $M_1$ to its corresponding parameter $\theta_2$ in $M_2$ (or $\phi$ if there is no such corresponding parameter).
As previously discussed in Section~\ref{mllms}, we only tackle the heterogeneous MLLMs that are designed from same pre-trained language model architecture, but adapts different modifications on model structure.

The principle of designing the mapping $f$ is that for the shared weights between the models (e.g. the weights in the pre-trained model), we can map them directly, and for additional weights in $M_1$, we map it to its original corresponding weight in $M_2$ if it is a duplicated multimodal parameter in $M_1$'s specific design, otherwise we map $\phi$ with it and apply no operation later in merging. In this way, we leverage the additional weights as much as possible without mapping irrelevant parameters together.


\subsection{Merging}
We follow the paradigm in Task Arithmetic \cite{task-arithmetic} to apply merging operation on task vectors and apply linear interpolation on them. 
Task vectors are defined as the fine-tuned parameters subtracted by pre-train weights: $\tau_i = \theta_i - \theta_0 \ (i=1,2)$, where $\theta_1$ and $\theta_2$ is the models to be merged, and $\theta_0$ is their common initialization point (e.g. the shared pre-trained weights for two different finetuned model).
Linear interpolation offers the availability to directly control the tendency between the two models alongside its simplicity. This allows us to actively adapt to different downstream tasks, as different downstream tasks often requires different combination or tendency on the two models for the best performance.
Linear interpolation on task vectors can be simply formatted as:  $\theta_{out} = \theta_0 + (1-\alpha) \tau_1 + \alpha \tau_2$, 
\noindent, where $\alpha$ is the linear interpolation coefficient. This is equivalent to: $\theta_{out} = (1-\alpha) \theta_1 + \alpha \theta_2$.
Thus, leveraging our mapping funciton $f$ in Section~\ref{mapping}, we can apply our merging operation on heterogeneous MLLMs as follow:
\begin{equation}
    \theta_{out}^i =\left\{
    \begin{aligned}
        &\theta_1^i, \quad & \mathsf{if}  f(\theta_1^i) = \phi
        \\
        &(1-\alpha) \theta_1^i + \alpha f(\theta_1^i),  \quad &          \mathsf{otherwise}
    \end{aligned} \right.
\end{equation}
Note that $f(\theta_1)$ is the parameter in $M_2$ that corresponds to $\theta_1$, according to the mapping function $f$. Now we can define the merging process of two model parameters $\Theta_1=\{\theta_1^i\}$ and $\Theta_2=\{\theta_2^j\}$ accordingly:
\begin{equation}
    \operatorname*{Merge}(\Theta_1, \Theta_2; f; \alpha) = \{\theta_{out}^i\} \label{eq:merge}
\end{equation}






\subsection{Searching} \label{searching}


Consider the base model, which is defined as the architecture that will be used after the merging process, containing $N$ scalar weight elements, then the merging process can be seen as the operation on a $N$-dimentional vector space $\mathbb{R}^N$, where the merging strategy $F$ transform two input points as initial parameters $\Theta_1$ and $\Theta_2$ to the merged parameters $\Theta_{out}$. For simplicity, we consider the case when the merging strategy $F$ takes one hyper-parameter $\alpha$ (linear interpolation coefficient).
\[
\Theta_{out} = F(\Theta_1, \Theta_2; \alpha), \quad \Theta_1, \Theta_2, \Theta_{out} \in \mathbb{R}^N
\]
Given the inputs $t_i$ on a downstream task, this creates a landscape on $\mathbb{R}^N$ that each model parameter corresponds to the model performance $S_{t_i}$ on the inputs.
\[
S_{t_i}(\Theta_{out}) = S_{t_i, F}(\Theta_1, \Theta_2, \alpha)
\]
Therefore, the goal of the hyper-parameter searching is to find the best $\alpha$ that maximize the merged model's performance on the tasks. For simplicity, we omit the F in the index.
\[
\alpha^* = \operatorname*{argmax}_{\alpha}(S_{t_i}(\Theta_1, \Theta_2, \alpha))
\]
Note that as the landscape varies greatly in different tasks, the best $\alpha$ may be different as well.

Previous model merging methods mainly relies on a validation set to search for the best hyper-parameter $\alpha$ with supervised searching. Formally, they use the best $\alpha$ in validation set $\hat{t_i}$ to approximate the best $\alpha$ in test set $t_i$.
\[
\hat{\alpha} = \operatorname*{argmax}_{\alpha}(S_{\mathbf{\hat{t_i}}}(\Theta_1, \Theta_2, \alpha)) \approx \operatorname*{argmax}_{\alpha}(S_{\mathbf{t_i}}(\Theta_1, \Theta_2, \alpha))
\]

However, this supervised way of searching has certain disadvantages: (1) labeled data with ground truth is hard to collect in some scenarios, and (2) the distribution shift between the validation set and the test set, even on the same task, will interfere the selection of the best hyper-parameter $\alpha^*$. To get rid of these drawbacks, we propose an unsupervised hyper-parameter selection method through a performance estimation metric that requires no labeled data.

\begin{figure}
    \centering
    \includegraphics[width=0.4 \textwidth, bb=0 0 461 346]{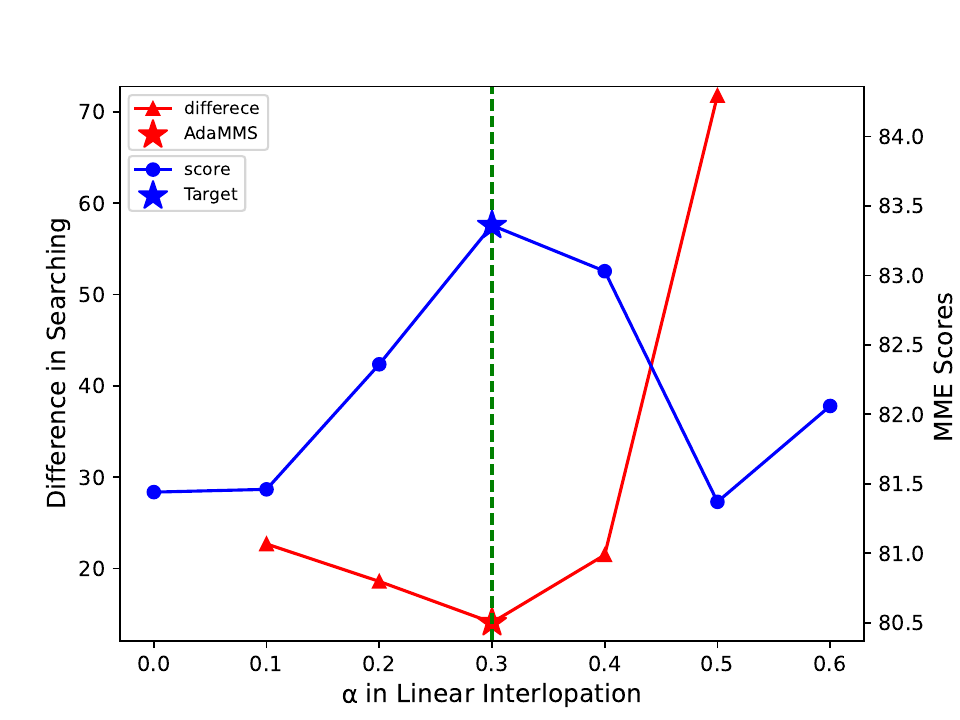}
    \caption{Results on merging LLaVA-v1.5-7B into Qwen2-VL-7B. The $\alpha$ with the best perfo, bb=0 0 461 346rmance are the same as the $\alpha$ with the fewest response differences.}
    \label{fig:difference}
\end{figure}

Specifically, we discover that the difference of generated responses between two adjacent $\alpha$ candidates can be used to estimate the model performance, and the best $\alpha$ can be approximated by the one with the lowest adjacent difference. 
As shown in Figure~\ref{fig:difference}, the trend of model performance is similar with its generation consistency which is measured by response differences, and the $\alpha$ with the highest generation consistency match the $\alpha$ with the highest performance.
Formally, let $\alpha^-$ and $\alpha^+$ be adjacent candidates on both sides of $\alpha$, respectively, and $D_{t_i}(\alpha; \alpha^-, \alpha^+)$ denoting the difference of generated responses between , we take:
\[
\bar{\alpha} = \operatorname*{argmin}_{\alpha}(D_{t_i}(\alpha; \alpha^-, \alpha^+)) \approx \operatorname*{argmax}_{\alpha}(S_{t_i}(\Theta_1, \Theta_2, \alpha))
\]
as the approximation of the best choice for $\alpha$. This eliminates the need of labeled data with ground truth, and avoids the data distribution shift between validation set and test set. 

The discovery indicates that near the best performance point, the model response tends to be more stable. This can be explained via a convex hypothesis. Suppose the landscape on given task $t_i$ is convex in a subspace of the parameter space that covers the candidate results of merged models, that is, for any $\lambda \in [0,1]$ we have:
\begin{equation*}
    S_{t_i}(\lambda\Theta_1+(1-\lambda)\Theta_2) \geq \
     S_{t_i}(\lambda\Theta_1) + \
      S_{t_i}((1-\lambda)\Theta_2)
\end{equation*}
This guarantee that the optimum is attained where the gradient vanishes:
\begin{equation*}
    \nabla_{\Theta}S_{t_i}(\Theta^*) = 0
\end{equation*}
where $\Theta^* \in \mathbb{R}^N$ is the optimal parameter value. In a convex function, the Hessian $H(\Theta^*)=\nabla^2_{\Theta}S_{t_i}(\Theta^*)$ at the optimum $\Theta^*$ is positive semi-definite, which implies local stability in a neighborhood around $\Theta^*$. The stability can be characterized by the second-order Taylor expansion:
\begin{equation*}
    S_{t_i}(\Theta) \approx S_{t_i}(\Theta^*) + \frac{1}{2}(\Theta-\Theta^*)^{\top}H(\Theta^*)(\Theta-\Theta^*)
\end{equation*}
Since $H(\Theta^*) \geq 0$, small deviations from $\Theta^*$ will result in small increases in the performance, ensuring relative stability on the landscape. Although the convex hypothesis is ideal, we confirm the effectiveness of our unsupervised hyper-parameter selection method in various experiments.
Proof in Appendix~\ref{appendix:proof} further shows the relationship between generation consistency and model performance.


We also find that while the landscape often changes between the validation and test sets and across different tasks, it remains consistent between the full test set and a small subset. This suggests that we can perform unsupervised hyper-parameter selection on a smaller subset of the data $\bar{t_i}$ without compromising its accuracy.
\[
\bar{\alpha} = \operatorname*{argmin}_{\alpha}(D_{\mathbf{\bar{t_i}}}(\alpha; \alpha^-, \alpha^+)) \approx \operatorname*{argmax}_{\alpha}(S_{\mathbf{t_i}}(\theta_1, \theta_2, \alpha))
\]

In conclusion, the process of \ours is described in Algorithm~\ref{alg:composition}.








\begin{algorithm}
\caption{\ours Procedure}
\label{alg:composition}
\begin{algorithmic}[1]
\Require Original MLLMs $M_1, M_2$, their parameters $\Theta_1, \Theta_2$ respectively, a subset of test inputs $\bar{t}$, and the candidates of the hyper-parameter $\{\alpha_n\}$
\Ensure Merged parameters $\Theta_{\text{out}}$ on $M_1$'s architecture

 \State Define a mapping $f$ that maps each parameter $\theta_1$ in $M_1$ architecture with its corresponding parameter $\theta_2$ in $M_2$ architecture (if exists) \Comment{Section~\ref{mapping}}
\State Define a process $\operatorname*{Generate}(\Theta, \bar{t})$ that returns the generation responses $G$ of model with parameters $\Theta$ on inputs $\bar{t}$
\State Define a function $\operatorname*{DiffCnt}(G_i, G_j)$ that counts the number of corresponding elements in $G_i$ and $G_j$ that do not exactly match
\For{$i = 1$ to $n$}
    \For{each hyper-parameter candidate $\alpha_i$ in $\{\alpha_n\}$}
        \State $\Theta_{cand}^{i} \gets \operatorname*{Merge}(\Theta_1, \Theta_2; f; \alpha_i)$ \Comment{Equation \eqref{eq:merge}}
        \State $G_i \gets \operatorname*{Generate}(\Theta_{cand}^{i}, \bar{t})$
    \EndFor
\EndFor
\For{$i = 2$ to $n-1$} \Comment{Assuming $\{\alpha_n\}$ is monotonic}
    \For{each hyper-parameter candidate $\alpha_i$ in $\{\alpha_n\}$}
        \State $D_i \gets \operatorname*{DiffCnt}(G_i, G_{i-1}) + \operatorname*{DiffCnt}(G_i, G_{i+1})$
    \EndFor
\EndFor
\State $i^* \gets \operatorname*{argmin}_{i}(D_i)$
\State $\Theta_{\text{out}} \gets \Theta_{\text{cand}}^{i^*}$
\State \Return $\Theta_{\text{out}}$
\end{algorithmic}
\end{algorithm}

\section{Experiment}

\label{sec:exp}
\subsection{Baselines}
Previous model merging methods cannot be directly applied to heterogeneous MLLMs with architecture difference, and our mapping method enables the fusion of heterogeneous models by transforming them into a homogeneous parameter space. Therefore, all the baseline experiments are conducted under the precondition of the mapping step of our proposed method. We consider the following model merging methods as our baselines:
\begin{itemize}

\item \textbf{Task Arithmetic} \cite{task-arithmetic} introduces the idea of task vectors and integrates them into the original pre-trained model for multi-task learning.

\item \textbf{Ties-Merging} \cite{ties} further addresses interferences in Task Arithmetic by removing unnecessary parameters from Task Arithmetic \cite{task-arithmetic}. This process eliminates redundant parameters and resolves symbol conflicts through Trim, Elect Sign, and Disjoint Merge steps.

\item \textbf{DARE} \cite{dare} tackles the parameter conflict problem in model merging by applying a drop and rescale operation before merging model weights. There are two variants of DARE: \textbf{Dare-Linear} and \textbf{Dare-Ties}, which perform different merging strategies after the drop and rescale operation. Dare-Linear performs linear interpolation, and Dare-ties performs Ties-Merging \cite{ties}.

\item \textbf{MetaGPT} \cite{metagpt} separate the data term and scaling coefficients in the optimization objective, which leads to a task-agnostic closed-form solution for the scaling coefficient.
    
\end{itemize}
\subsection{Models} \label{models}

We have conducted extensive experiments on the combinations of existing open-source 7B-scale MLLMs. 
Since most of the top-performing open-source MLLMs are currently based on two language model architectures, Qwen2 \cite{qwen2} and LLaMA \cite{llama}, we selected representative and outstanding MLLMs derived from each model for our main experiments.
Specifically, on Qwen2 architecture, we merge LLaVA-OneVision-Qwen-7B \cite{llava-onevison} into Qwen2-VL-7B \cite{qwen2-vl}, and on LLaMA architecture, we merge LLaVA-v1.5-7B \cite{llava1.5} into CogVLM-Chat-7B \cite{cogvlm}.

We have also conducted experiments on combinations of LLaMA-based MLLMs, including combinations LLaVA-v1.5-7B, CogVLM-Chat-7B, ShareGPT4V-7B \cite{sharegpt4v} and mPLUG-Owl2-LLaMA2-7B \cite{mplugowl2}.
See Appendix A for more details. 

\begin{table*}[!ht]
    \centering
    \resizebox{\textwidth}{!}{%
    \begin{tabular}{lllllllllcc}
        \toprule             

        Model & $\mathrm{MMMU_{val}}$ &  $\mathrm{MME_{sum}}$ &  $\mathrm{SeedBench_{all}}$ & $\mathrm{OCRBench}$  &  $\mathrm{TextVQA_{val}}$  & $\mathrm{OKVQA}$ & $\mathrm{GQA}$  &  $\mathrm{VizWiz_{val}}$ & $\mathrm{SUM}$  & $\mathrm{Top2}$  \\ 
        
        \hline
\rowcolor{gray!20}
\multicolumn{11}{c}{\textbf{Original Models}} \\
\hline
        Qwen2-VL\footnotesize(base) & 50.11 & 81.44 & 75.85 & \textit{86.00} & \textit{84.12} & 51.43 & 61.80 & 68.32 & 559.07 & 2 \\ 
        LLaVA-OneVision  & 43.44 & 77.04 & 75.44 & 69.60 & 78.47 & 49.57 & 59.84 & 60.97 & 514.37 & 0 \\ 
     \hline
       
\rowcolor{gray!20}
\multicolumn{11}{c}{\textbf{Baselines}} \\
\hline

        Task Arithmetic & 
        48.44\footnotesize(+1.67) & 82.33\footnotesize(+3.09) & 75.81\footnotesize(+0.17) & 77.90\footnotesize(+0.10) & 76.22\footnotesize(-5.08) & 50.60\footnotesize(+0.10) & \textbf{62.26\footnotesize(+1.44)} & 62.76\footnotesize(-1.89) & 536.32\footnotesize(-0.40) &1 
 \\ 
        
        Ties-Merging & \textbf{51.11\footnotesize(+4.34)} & \underline{82.65\footnotesize(+3.41)} & \underline{76.29\footnotesize(+0.64)} & 84.40\footnotesize(+6.60) & 79.56\footnotesize(-1.74) & \underline{52.56\footnotesize(+2.06)} & 61.84\footnotesize(+1.02) & 66.34\footnotesize(+1.69) & 554.75\footnotesize(+18.03) &4 \\ 
        
        DARE-Linear & 43.78\footnotesize(-3.00) & 66.06\footnotesize(-13.18) & 74.32\footnotesize(-1.33) & 72.40\footnotesize(-5.40) & 64.65\footnotesize(-16.65) & 43.41 \footnotesize(-7.09) & 55.13\footnotesize(-5.69) & 50.18\footnotesize(-14.47) & 469.93\footnotesize(-66.79) & 0 \\ 
        
        DARE-Ties & 45.00\footnotesize(-1.78) & 54.43\footnotesize(-24.81) & 74.07\footnotesize(-1.58) & 75.20\footnotesize(-2.60) & 78.54\footnotesize(-2.76) & 49.61\footnotesize(-0.89) & 58.51\footnotesize(-2.31) & 58.05\footnotesize(-6.60) & 493.41 \footnotesize(-43.31) & 0 \\ 
        
        MetaGPT & 50.67\footnotesize(+3.90) & 81.21\footnotesize(+1.97) & \textbf{76.35\footnotesize(+0.70)} & \textbf{85.50\footnotesize(+7.70)} & \underline{83.63\footnotesize(+2.33)} & 52.24\footnotesize(+1.74) & 61.99\footnotesize(+1.17) & \textbf{69.16\footnotesize(+4.51)} & \underline{560.75\footnotesize(+24.03)} & 5  \\[0.5ex] 
        
       \hline
       
\rowcolor{gray!20}
\multicolumn{11}{c}{\textbf{Our Method}} \\
\hline

        \ours & 
        \textbf{51.11\footnotesize(+4.34)} & \textbf{83.36\footnotesize(+4.12)} & 76.20\footnotesize(+0.55) & \textbf{85.50\footnotesize(+7.70)} & \underline{83.41\footnotesize(+2.11)} & \textbf{53.56\footnotesize(+3.06)} & \underline{62.02\footnotesize(+1.20)} & \underline{68.40\footnotesize(+3.75)} & \textbf{563.56\footnotesize(+26.84)} & 8 \\ 
        \bottomrule
    \end{tabular}%
        }
    \caption{Results on merging LLaVA-OneVision-7B into Qwen2-VL-7B. All the scores have been scaled to 0-100. SUM refers to the sum of scores on all tasks after scaling. Top2 column represents the number of tasks obtained by this method from the top two among all methods. The number in the parenthesis indicates the performance improvement compared with the average score of original models. The results in the original models that are higher than all model merging methods are highlighted in italics.}
    \label{tab:llava2qwen}
\end{table*}

\begin{table*}[!ht]
    \centering
    \resizebox{\textwidth}{!}{%
    \begin{tabular}{lllllllllcc}
    \toprule
        {Model} & $\mathrm{MMMU_{val}}$ &  $\mathrm{MME_{sum}}$ &  $\mathrm{SeedBench_{all}}$ & $\mathrm{OCRBench}$  &  $\mathrm{TextVQA_{val}}$  & $\mathrm{OKVQA}$ & $\mathrm{GQA}$  &  $\mathrm{VizWiz_{val}}$ & $\mathrm{SUM}$  & $\mathrm{Top2}$  \\ 

\hline
\rowcolor{gray!20}
\multicolumn{11}{c}{\textbf{Original Models}} \\
\hline
        
        CogVLM\footnotesize(base) & 34.80  & 59.23  & 61.22  & \textit{56.50}  & \textit{77.57}  & 60.82  & 59.43  & {37.09}  & {446.66 } &2 \\ 
        LLaVA & 35.10 & 66.68 & 60.52 & 31.30 & 46.04 & 53.42 & \textit{61.94 }& 54.29 & 409.29  &0
 \\
\hline
\rowcolor{gray!20}
\multicolumn{11}{c}{\textbf{Baselines}} \\
\hline
        
        Task Arithmetic & \underline{36.20  \footnotesize(+1.25)} & \underline{65.99}  \footnotesize(+3.03) & \textbf{65.85  \footnotesize(+4.98) }& 51.20  \footnotesize(+7.30) & 68.21  \footnotesize(+6.40) & \textbf{61.92  \footnotesize(+4.80)} & 58.82  \footnotesize(-1.87) & 35.70  \footnotesize(-9.99) & 443.89 
 \footnotesize(+15.91) & 4  \\

        Ties-Merging & 34.00  \footnotesize(-0.95) & 57.29  \footnotesize(-5.67) & 38.97  \footnotesize(-21.90) & 55.00  \footnotesize(+11.10) & 59.73  \footnotesize(-2.08) & 40.31  \footnotesize(-16.81) & 51.97  \footnotesize(-8.72) & 24.36  \footnotesize(-21.33) & 361.63 
  \footnotesize(-66.35) & 0 \\

        DARE-Linear & \textbf{36.80  \footnotesize(+1.85)} & 64.08  \footnotesize(+1.12) & \underline{65.07  \footnotesize(+4.20)} & 47.90  \footnotesize(+4.00) & 65.35  \footnotesize(+3.54) & 60.96  \footnotesize(+3.84) & 58.01  \footnotesize(-2.68) & 36.12  \footnotesize(-9.57) & 434.29 
  \footnotesize(+6.31) & 2 \\

        DARE-Ties & 33.60  \footnotesize(-1.35) & 46.75  \footnotesize(-16.21) & 58.41  \footnotesize(-2.46) & 26.50  \footnotesize(-17.40) & 50.48  \footnotesize(-11.33) & 53.15  \footnotesize(-3.97) & 49.62  \footnotesize(-11.07) & 31.43  \footnotesize(-14.26) & 349.94 
  \footnotesize(-78.04) & 0  \\

        MetaGPT & 34.70  \footnotesize(-0.25) & 59.37  \footnotesize(-3.59) & 61.29  \footnotesize(+0.42) & \textbf{56.40  \footnotesize(+12.50)} & \textbf{76.96  \footnotesize(+15.15)} & 60.84  \footnotesize(+3.72) & \underline{59.44  \footnotesize(-1.25)} & \underline{36.97  \footnotesize(-8.72)} & \underline{445.97 
 \footnotesize(+17.99)} & 5  \\[0.5ex] 
        
\hline 
\rowcolor{gray!20}
\multicolumn{11}{c}{\textbf{Our Method}} \\
\hline 
      
        AdaMMS & 34.90  \footnotesize(-0.05) & \textbf{69.09  \footnotesize(+6.13)} & 64.12  \footnotesize(+3.25) & \underline{55.70  \footnotesize(+11.80)} & \underline{76.90  \footnotesize(+15.09)} & \underline{61.11  \footnotesize(+3.99)} & \textbf{60.12  \footnotesize(-0.57)} & \textbf{37.27  \footnotesize(-8.42) }& \textbf{459.21  \footnotesize(+31.23)} & 7  \\ 
        
        \bottomrule
    \end{tabular}%
        }
    \caption{Results on merging LLaVA-v1.5-7B into CogVLM-chat-7B.}
     \label{tab:cogvlm}

\end{table*}



\subsection{Benchmarks}
To evaluate the capabilities of the merged MLLMs, we have conducted experiments on various benchmarks that cover a wide range of vision-language abilities. According to the classification in \cite{survey_benchmark}, our benchmarks fall into three categories: (1) comprehensive-evaluation, (2) cognition and reasoning, (3) text-rich VQA. 
The comprehensive-evaluation tasks consist of MME \cite{mme}, SeedBench \cite{seedbench} and VizWiz \cite{vizwiz}. Cognition and reasoning type include MMMU \cite{mmmu}, OK-VQA \cite{okvqa} and GQA \cite{gqa}. Text-rich VQA type encompass  OCRBench \cite{ocrbench} and TextVQA \cite{textvqa}.

To present the overall performance of the MLLMs in a standardized manner, we apply linearly normalization to the scores across all tasks, scaling them to a range from 0 to 100. Specially, the total score of MME is 2800, which we have divided by 28 for scaling purposes.

\subsection{Implementation} \label{impl}

To apply our unsupervised hyper-parameter selection method in the searching step, we need to specify the candidates of $\alpha$ (linear interpolation coefficient). We sample the candidates in a subinterval of $[0,1]$ with a fixed granularity.

\noindent\textbf{Subinterval} \ We find that merging with $\alpha\geq0.7$ often results in collapsing language ability of the merged model, therefore we empirically limit the subinterval of $\alpha$ candidates to $[0,0.6]$ for eliminating unnecessary search.


\noindent\textbf{Granularity} \  
We use granularity to determine the interval between two adjacent candidates of $\alpha$. In our main experiments, we choose the granularity as 0.1 to obtain satisfying performance with acceptable computation cost.
\noindent\textbf{Evaluation Framework} \ We evaluated the benchmarks with LMMs-Eval \cite{lmms-eval} and VLMEvalKit \cite{vlmevalkit}, two open-source evaluation frameworks for MLLMs.

\noindent\textbf{Subset Searching} \ Instead of conducting search across the entire available input data, we strategically utilize a small subset of only 100 inputs during the search phase, reducing the data volume by at least an order of magnitude. Experimental results demonstrate that this maintains performance without compromising effectiveness.

\section{Results}
\label{sec:results}

\begin{table*}[!ht]
    \centering
    \resizebox{\textwidth}{!}{%
    \begin{tabular}{lllllllllll}
    \toprule
        Method & $\mathrm{MMMU_{val}}$ &  $\mathrm{MME_{sum}}$ &  $\mathrm{SeedBench_{all}}$ & $\mathrm{OCRBench}$  &  $\mathrm{TextVQA_{val}}$ & $\mathrm{ScienceQA}$ & $\mathrm{OKVQA}$ & $\mathrm{GQA}$  &  $\mathrm{VizWiz_{val}}$ & $\mathrm{SUM}$ \\ 
        \midrule

         EM-Full & 51.11 & 83.36 & 76.34 & 85.50  & 83.41 & 85.69 & 53.56 & 62.02 & 68.40  & 563.70   \\ 
         Emb-Full & 50.56 & 83.36 & 76.34 & 85.50  & 83.41 & 85.69 & 53.56 & 61.44 & 68.40  & 562.57   \\ 
         EM-Sample100 & 51.11 & 83.36 & 76.20  & 85.50  & 83.41 & 86.55  & 53.56 & 62.02 & 68.40  & 562.12  \\   
         Emb-Sample100 & 51.11 & 82.36 & 76.34 & 85.50  & 83.41 & 85.69 & 53.56 & 61.44 & 68.40  & 563.56  \\

        \bottomrule
    \end{tabular}%
        }
    \caption{Results on AdaMMS when merging LLaVA-OneVison-7B into Qwen2-VL-7B using exact match (EM-) and sentence embedding (Emb-) to calculate the differences in searching phase, using full test set inputs (-Full) and a sampled subset of 100 inputs (-Sample100).}
     \label{tab:embedding}

\end{table*}

\begin{table}[ht]
    \centering
    \begin{tabular}{lccc}
        \toprule
        {Model} & $\mathrm{MMMU_{val}}$ &  $\mathrm{MME_{sum}}$ & $\mathrm{OCRBench}$   \\ 
        \midrule
        \(\alpha\)-0.00 & 50.11  & 81.44  & \textbf{86.00}  \\ 
       
        \(\alpha\)-0.10 & 50.56  & 81.46  & 85.50  \\ 
        
        \(\alpha\)-0.20 & 51.11  & 82.36  & 85.20  \\ 
       
    \(\alpha\)-0.30 & \textbf{51.22}  & \textbf{83.36}  & 84.40  \\ 
     
        \(\alpha\)-0.40 & 50.67  & 83.03  & 80.70  \\ 
     
        \(\alpha\)-0.50 & 50.00  & 81.37  & 76.40  \\ 
     
        \(\alpha\)-0.60 & 47.00  & 82.06  & 71.20 \\
          \midrule
    Oracle & 51.22\footnotesize(0.30) & 83.36\footnotesize(0.30) & 85.50\footnotesize(0.10) \\   
       \ours &   51.11\footnotesize(0.20)  & 83.36\footnotesize(0.30)  &  85.50\footnotesize(0.10) \\
     \bottomrule
    \end{tabular}%
    \caption{Results with $\alpha$ granularity of 0.1 when merging LLaVA-OneVision-7B into Qwen2-VL-7B. The values in parentheses indicate the selected $\alpha$. Oracle represents the best possible performance (upper bound) for each task, while AdaMMS shows the results achieved by our unsupervised selection method.}
    \label{tab:qwen0.1}
\end{table}

\begin{table}[!ht]
    \centering
     \resizebox{0.4 \textwidth}{!}{%
    \begin{tabular}{lcc}
    \\
    \hline
   
        \rowcolor{gray!20}
\multicolumn{3}{c}{\textbf{Original Models}} \\
\hline 

        LLaVA-OneVision & 69.60  & 69.60  \\ 
        Qwen2-VL & 86.00  & 86.00  \\ 

          \hline

        Merging-Base & LLaVA-OneVision & Qwen2-VL \\ 
        
         \hline       
        \rowcolor{gray!20}
        
\multicolumn{3}{c}{\textbf{Baselines}} \\
\hline 
        Task Arithmetic  & 68.10  & 77.90  \\ 
        Ties-Merging  & 56.10  & 84.40  \\ 
        DARE-Linear  & 63.90  & 72.40  \\ 
        DARE-Ties  & 64.40  & 75.20  \\ 
        MetaGPT & 38.40  & \textbf{85.50}  \\ 
         \hline       
        \rowcolor{gray!20}
\multicolumn{3}{c}{\textbf{Linear Interpolation}} \\
\hline 
        $\alpha$-0.10 & 70.60  & 85.50  \\ 
        $\alpha$-0.20 & 71.70  & 85.20  \\ 
        $\alpha$-0.30 & 69.90  & 84.40  \\ 
        $\alpha$-0.40 & 67.00  & 80.70  \\ 
        $\alpha$-0.50 & 62.00  & 76.40  \\ 
        $\alpha$-0.60 & 54.40  & 71.30  \\ 
 \hline       
        \rowcolor{gray!20}
\multicolumn{3}{c}{\textbf{Our Method}} \\
\hline 
              
        AdaMMS & \textbf{70.60}  & \textbf{85.50} \\ \hline
    \end{tabular}
    }
    
    \caption{Results on OCRBench when merging LLaVA-OneVision-7B and Qwen2-VL-7B. }
    \label{tab:reverse}
\end{table}

As described in Section~\ref{models}, the main results on distinct vision-language benchmarks are conducted with two representative MLLM pairs. Specifically, Table~\ref{tab:llava2qwen} shows the results of merging LLaVA-OneVision-7B's parameters into Qwen2-VL-7B's parameters and architecture, and Table~\ref{tab:cogvlm} shows the results of merging LLaVA-v1.5-7B's parameters into CogVLM-chat-7B's parameters and architecture. Results of other model pairs and larger models can be found in Appendix~\ref{appendix:additional}.

\textbf{\ours  addresses the challenges of merging for heterogeneous MLLMs and outperforms strong baselines.} As demonstrated in Table~\ref{tab:llava2qwen} and Table~\ref{tab:cogvlm}, our proposed \ours model merging method achieves the highest cumulative performance scores across both MLLM pairs, indicating its effectiveness in merging heterogeneous MLLMs. Ranks among the top two performs in 8 out of 9 metrics in Table~\ref{tab:llava2qwen} and 7 metrics in Table~\ref{tab:cogvlm}, demonstrating its consistent ability to adaptively improve performance across most tasks. 
Moreover, our method stands out as the only approach where the merged model significantly outperforms both pre-merged models, achieving an average gain of +3.36~(total gain of +26.84) over Qwen2-VL and +3.90 over CogVLM across 8 tasks.
Given that most baseline methods employ supervised search techniques that incorporate additional information, our unsupervised search approach demonstrates exceptional performance on vision-language benchmarks, as detailed illustrated in Appendix~\ref{appendix:unsupervised}. 

\textbf{The proposed unsupervised hyper-parameter selection method is able to select a near-optimal $\alpha$.} We evaluate the performance across different coefficient values $\alpha$, comparing the results obtained through our unsupervised hyper-parameter selection method against those achieved with the optimal $\alpha$ chosen by the actual best results, which serves as the theoretical upper bound. As shown in Table \ref{tab:qwen0.1}, our method consistently performs remarkably close to this upper bound, with a maximum deviation of only 0.5 points. These results demonstrate the capability of our method to accurately identify near-optimal $\alpha$ values, achieving performance levels approaching the theoretical best.


Note that on the OCRBench and TextVQA benchmarks, all model merging methods, including \ours, show a performance drop compared to the original base model.
We hypothesize that this is due to the large performance gap between the two original models on these benchmarks.
Even though, \ours still outperforms most of the baselines, showing the robustness of our method on various scenarios.

\section{Analysis}
\label{sec:ana}

\subsection{Different Factors for Calculating Generation Consistency}
We conducted analytical experiments on our generation consistency calculation methods, focusing on two key factors: the choice between using a 100-sample subset versus the complete dataset, and the selection of evaluation metrics. In the searching step of our method, we employed an exact match metric to calculate $\operatorname*{DiffCnt}$ in Algorithm~\ref{alg:composition}, which serves as our generation consistency indicator for model performance prediction. Given that exact match is a binary, rigorous evaluation metric, we explored an alternative, more flexible approach to measure generation consistency. Specifically, we computed the cosine similarity between sentence embeddings generated by all-MiniLM-L6-v2 \cite{sentence-bert}, which was used to calculate $\operatorname*{DiffCnt}$. The analysis results are presented in Table~\ref{tab:embedding}. Although embeddings theoretically offer more fine-grained semantic representations, our results demonstrate that the embedding-based metric performs comparably to the exact match metric. Furthermore, our experiments confirm that sampling 100 instances achieves results nearly equivalent to the complete dataset.


\subsection{Merging with Large Performance Gap}

As discussed in Section~\ref{sec:results}, all model merging methods experience performance drop after the merging on two benchmarks, OCRBench and TextVQA. It shows that merging a model with significant \textit{lower} performance into the base model will decrease the performance on the task. Conversely, in Table~\ref{tab:reverse}, the merging from Qwen2-VL-7B to LLaVA-OneVision-7B shows that merging a model with significant \textit{higher} performance into the base model will not necessarily improve the model performance. And in this case, \ours is the only model merging method that resists the performance drop after merging.
In general, we observed that original models with similar performance tends to benefit from model merging, while original models with large performance gap do not.

\begin{figure}
    \centering
    \includegraphics[width=\linewidth, bb=0 0 678 389]{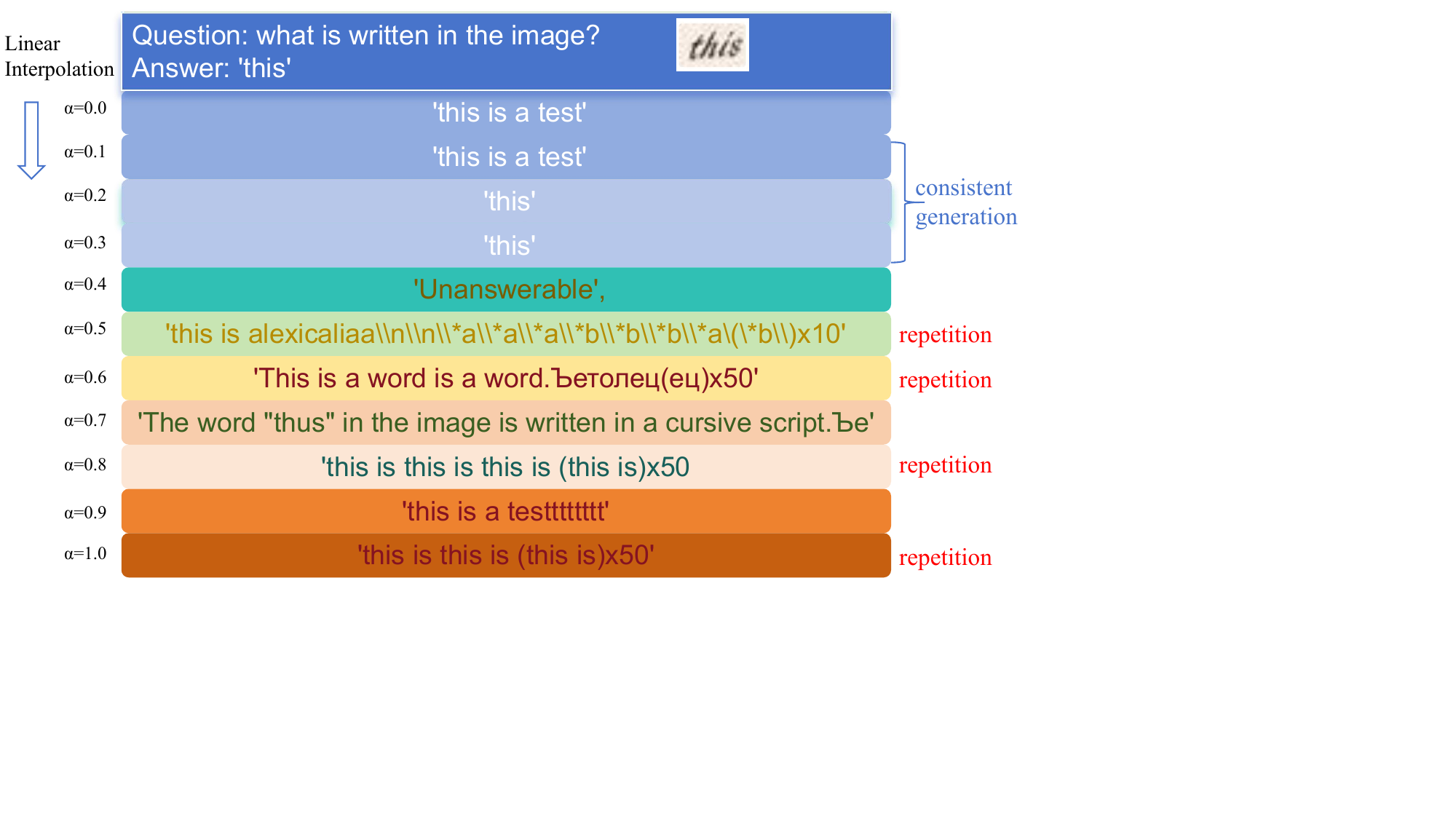}
    \caption{Model responses with the change of $\alpha$ in linear interpolation. Similar colors indicate similar responses.  }
    \label{fig:asymmetry}
\end{figure}

\subsection{Asymmetry in the Parameter Space of Heterogeneous Models} \label{sec:asymmetry_sec}
In Section~\ref{impl}, we discussed that merging with large $\alpha$ often results in collapsing language ability. To validate our choice of the subinterval $[0, 0.6]$ in determining candidates of $\alpha$, we demonstrate this phenomenon in Figure~\ref{fig:asymmetry}, which shows that the model generates consistently near the parameters of the base model with small $\alpha$, and collapses gradually with larger $\alpha$. We attribute the phenomenon to the asymmetry in the parameter space, as the two original models have unequal status that comes from the choice of base architecture.

\subsection{Selection of Granularity for \texorpdfstring{$\alpha$}{alpha}}
\label{sec:interval}
To validate our choice of the granularity in Section~\ref{impl}, we conducted additional experiments with various granularities of $\alpha$ candidates on n MME and OCRBench when merging LLaVA-OneVison-7B into Qwen2-VL-7B. As shown in Appendix~\ref{appendix:granularity}, the result shows that the difference of selected $\alpha$ and model performance do not change significantly with different granularities. This shows that our choice of granularity as 0.1 would result in comparable performance, with less computation cost.





\section{Conclusion}
\label{sec:conclusion}

In this work, we propose a novel model merging method \ours to address the challenges in merging heterogeneous MLLMs. We first connect the parameters of different MLLMs through a mapping function, enabling merging operations. We then apply linear interpolation to the mapped model weights to adaptively optimize performance across tasks. To optimize the interpolation coefficient without labeled data, we introduce an unsupervised hyperparameter searching method based on our discovery in the parameter space: model performance can be estimated through the generation consistency. We demonstrate that 100 data samples are enough to search for near-optimal coefficients effectively.
Extensive experimental results show that \ours outperforms existing model merging methods for MLLMs and successfully addresses the challenges in merging heterogeneous MLLMs. We hope that our work mitigates the limitations of heterogeneous model merging methods and provides valuable insights for future research on unsupervised performance estimation and optimization.


\section*{Acknowledgment}
    
This work is supported by the National Key R\&D Program of China (2022ZD0160502) and the National Natural Science Foundation of China (No. 62276152).

{
    \small
    \bibliographystyle{ieeenat_fullname}
    \bibliography{main}
}

\clearpage
\appendix


\section{Results on Additional Model Pairs}
\label{appendix:additional}

We conducted experiments on additional model pairs, summarized in Table~\ref{tab:avg6}, which highlights the cumulative performance gains across tasks for six different model pairs. The model pairs include: (1) merging LLaVA-OneVision \cite{llava-onevison} into Qwen2-VL \cite{qwen2-vl} (Table~\ref{tab:llava2qwen}), (2) merging LLaVA-v1.5 \cite{llava1.5} into CogVLM \cite{cogvlm} (Table~\ref{tab:cogvlm}), (3) merging mPLUG-Owl2 into LLaVA-v1.5, (4) merging LLaVA-v1.5 into mPLUG-Owl2 \cite{mplugowl2} (Table~\ref{tab:llava2mplug}), (5) merging CogVLM into mPLUG-Owl2 (Table~\ref{tab:cog2mplug}) and (6) merging mPLUG-Owl2 into CogVLM (Table~\ref{tab:mplug2cog}).
The performance gain for each task is computed as the difference between the performance of our method (or baselines) and the average performance of the two original models, with positive values indicating an improvement.
In Table~\ref{tab:avg6}, the SUM column presents the total performance gains across all tasks, where \ours outperforms all baselines, achieving +91.92 performance gain, and consistently ranks among the top two in performance gains across all benchmarks.
It is noteworthy that on GQA \cite{gqa} and VizWiz \cite{vizwiz} benchmarks in Table~\ref{tab:llava2mplug} and Table~\ref{tab:cog2mplug}, all model merging methods experience a performance drop. We attribute this decline to the significant performance gap between the original models on these benchmarks. In these scenarios, \ours demonstrates the smallest performance decrease among them.
In Table~\ref{tab:llava2mplug}, Table~\ref{tab:cog2mplug} and Table~\ref{tab:mplug2cog}, \ours obtains the second best result in the sum of all benchmarks, with a small gap compared to the best baseline. 

To investigate the effect of altering base models on performances, we analyze experiments on merging the same model pair with different base models. For the model pair of mPLUG-OWl2 and CogVLM, results in Table~\ref{tab:cog2mplug} use mPLUG-Owl2 as the base model, and results in Table~\ref{tab:mplug2cog} use CogVLM as the base model.
On benchmarks where the original models exhibit a significant performance gap, such as OCRBench \cite{ocrbench} and TextVQA \cite{textvqa}, model merging methods, including \ours, achieve only marginal performance improvements. In contrast, on benchmarks where the original models have comparable performance, \ours consistently enhances the base model’s performance (with the exception of GQA \cite{gqa} for the mPLUG-Owl2 architecture), irrespective of the choice of base model.
Notably, even when merging a weaker model into a stronger one for a specific task, \ours can sometimes boost the stronger model's performance. For instance, this effect is observed on SEEDBench \cite{seedbench}, OKVQA \cite{okvqa}, and GQA \cite{gqa} in Table~\ref{tab:mplug2cog}. These results highlight that our model merging technique can further optimize the performance of a strong model, even when another model demonstrates weaker performance on the same task.

Additionally, to demonstrate the effectiveness of our method on larger models, we conducted experiments on Cambrian and Yi-VL with 34B language model size. Table~\ref{tab:larger} shows that \textbf{\ours also merges the abilities in larger MLLMs effectively}.

\begin{table}[!ht]
    \centering\small
    \begin{tabular}{lll}
    \hline
        Model & $\mathrm{OCRBench}$ & $\mathrm{MME}$  \\ \hline
        Cambrian\footnotesize(base) & 58.70 & 72.50 \\ 
        Yi-VL & 29.70 & 73.65\\ 
        AVG & 44.20 & 73.08  \\        
        AdaMMS & \textbf{59.20} & \textbf{74.07}  \\ \hline
    \end{tabular}
    \caption{Results on merging Yi-VL into Cambrian.}
    \label{tab:larger}
\end{table}

\section{Implementation Details of \ours}
\label{appendix:impl}

The implementation details of \ours are as follows:

\noindent\textbf{Mapping} \  In this step, we identify parameters in the language models that account for additional weights. For CogVLM \cite{cogvlm}, all weights within the visual experts in the attention mechanism (including the QKV matrix and the FFN of the visual expert) are treated as additional weights. For mPLUG-Owl2 \cite{mplugowl2}, vision representation weights within the Modality-Adaptive Modules (such as the decoupled vision layer-norm and KV matrix) are considered additional weights. For different vision encoders, the vision encoder weights of the base model are retained as the final weights after merging, regardless of the vision encoder in the other model.

\noindent\textbf{Merging} \  During this step, we first merge the weights in the language model of the base model. If the weights are not classified as additional weights in the Mapping step, they are merged using linear interpolation or other baseline merging techniques. For weights categorized as additional weights, we check whether the other model has duplicated the same weights. Based on this, we (1) merge the weights if duplicates exist, or (2) retain the original weights in the base model if no duplicates are found.

\noindent\textbf{Searching} \  In the final step, we randomly select a subset of 100 test inputs to determine the optimal $\alpha$. For each $\alpha$ candidate, we generate model responses for the selected inputs. To select the best $\alpha$, we apply the Exact Match metric for the total difference score: for each input, if the merged model's response with a given $\alpha$ matches the response with adjacent $\alpha$ values, the difference score is 0; otherwise, it is 1. The total difference score is the sum of scores across all inputs in the subset. The $\alpha$ with the lowest total difference score is selected as the final choice. Note that the small subset of 100 inputs is randomly sampled using the method in LMMs-Eval framework \cite{lmms-eval}. We have repeated the sampling process to ensure that the randomness in sampling does not affect the performance of our method.

\section{Evaluation Details}
\label{appendix:eval}

We utilize LMMs-Eval \cite{lmms-eval} and VLMEvalKit \cite{vlmevalkit}, two open-source evaluation frameworks for MLLMs, to assess our models. Specifically, for evaluating MMMU \cite{mmmu}, MME \cite{mme}, SEEDBench \cite{seedbench}, OCRBench \cite{ocrbench}, and TextVQA \cite{textvqa} within the Qwen2-VL \cite{qwen2-vl} architecture, we use the VLMEvalKit framework, while LMMs-Eval is employed for the others. To ensure consistency with the reported results for LLaVA and mPLUG-Owl2 on OK-VQA \cite{okvqa}, we adapted the prompt template in the evaluation framework, as detailed in Table~\ref{tab:prompt}. Other prompt templates remains the same in the evaluation frameworks.

\section{Comparing Supervised and Unsupervised}
\label{appendix:unsupervised}

We compared \ours with baseline merging methods with supervised hyper-parameter selection. Due to the absence of separate test sets, we trained the supervised baseline on either a subset or the entirety of the evaluation set. \textbf{This implies that the supervised baseline was in a more favorable position compared to our method, as our method does not have access to the groudtruth labels.} Table~\ref{tab:supervised} shows that \ours still outperforms it, indicating the superiority of our unsupervised method.



\section{Intermediate Results in Searching}
\label{appendix:searching}

We present an example of the intermediate results during the selection of $\alpha$. As shown in Figure~\ref{tab:inter_alpha}, \ours effectively identifies a near-optimal $\alpha$, achieving performance close to the best possible outcome. Specifically, our unsupervised hyper-parameter selection method successfully chooses the optimal $\alpha$ candidate in half of the benchmarks and maintains a deviation of no more than 0.2 from the best $\alpha$ in the remaining cases.

Figure~\ref{fig:consistency_acc4} illustrates the relationship between model performance and generation consistency across MMMU, MME, SeedBench, and OCRBench when merging LLaVA-OneVision into Qwen2-VL. The observed trends validate our approach in the search step, where model performance is approximated using generation consistency without relying on labeled data. Notably, for these tasks, the $\alpha$ selected by our method corresponding to the highest generation consistency deviates from the $\alpha$ achieving the best performance by no more than 0.1, showing that our hyper-parameter selection method achieves near-optimal performance.

\begin{table}[h]
   
     \resizebox{0.478\textwidth}{!}{%
    \begin{tabular}{ccc}
    \hline
        Framework & Base Model & Prompt   \\ \hline
        \multirow{2}{*}{LMMs-Eval} & LLaVA &  \multirow{2}{*}{Answer the question using a single word or phrase.} \\
         & mPLUG-Owl2 &  \\ \hline
          
    \end{tabular}
    }
    \caption{Altered prompt for evaluation on OK-VQA.}
    \label{tab:prompt}
\end{table}

\section{Supplementary Proof}
\label{appendix:proof}

We provide the following proof as the theoretical justification for relationship between generation consistency correlates and model performance.

\noindent\textit{Proof.} Using the notation in Section~\ref{searching}, for an arbitrary task $t_i$, let $S_{t_i}(\alpha)$ be the ratio of correct answer at position $\alpha$, and $D_{t_i}(\alpha; \alpha^-)$ be the ratio of the difference in generated responses between position $\alpha$ and its adjacent candidate $\alpha^-$.
Since the difference in $S_{t_i}(\alpha)$ is only influenced  by the subset of generated responses where the correctness status changes (i.e., transitions between correct and incorrect), we have $|S_{t_i}(\alpha)-S_{t_i}(\alpha^-)| \leq D_{t_i}(\alpha; \alpha^-)$. For the same reason with $\alpha^+$, we can prove  $|S_{t_i}(\alpha)-S_{t_i}(\alpha^-)| + |S_{t_i}(\alpha)-S_{t_i}(\alpha^+)| \leq 2D_{t_i}(\alpha; \alpha^-, \alpha^+)$.
Therefore, a higher generation consistency with small $D_{t_i}(\alpha; \alpha^-, \alpha^+)$ implies a higher model performance $S_{t_i}(\alpha)$, due to its convexity.

\section{Experimental Results in Granularity for \texorpdfstring{$\alpha$}{alpha}}
\label{appendix:granularity}

\begin{figure*}
    \centering
   
    \begin{subfigure}[b]{0.43\textwidth}
        \centering
        \includegraphics[width=\textwidth, bb=0 0 720 432]{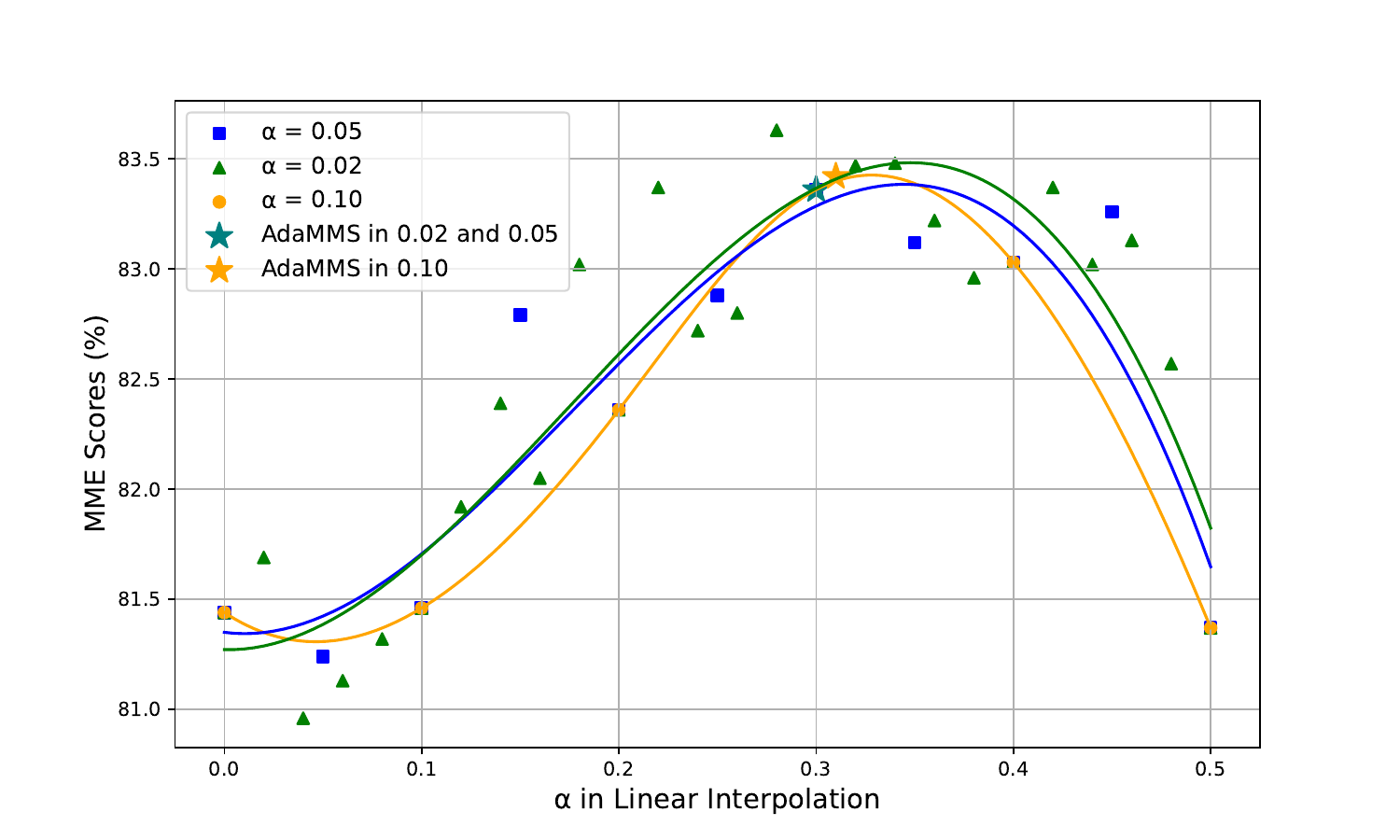}
    \end{subfigure}
    \begin{subfigure}[b]{0.43\textwidth}
        \centering
        \includegraphics[width=\textwidth, bb=0 0 720 432]{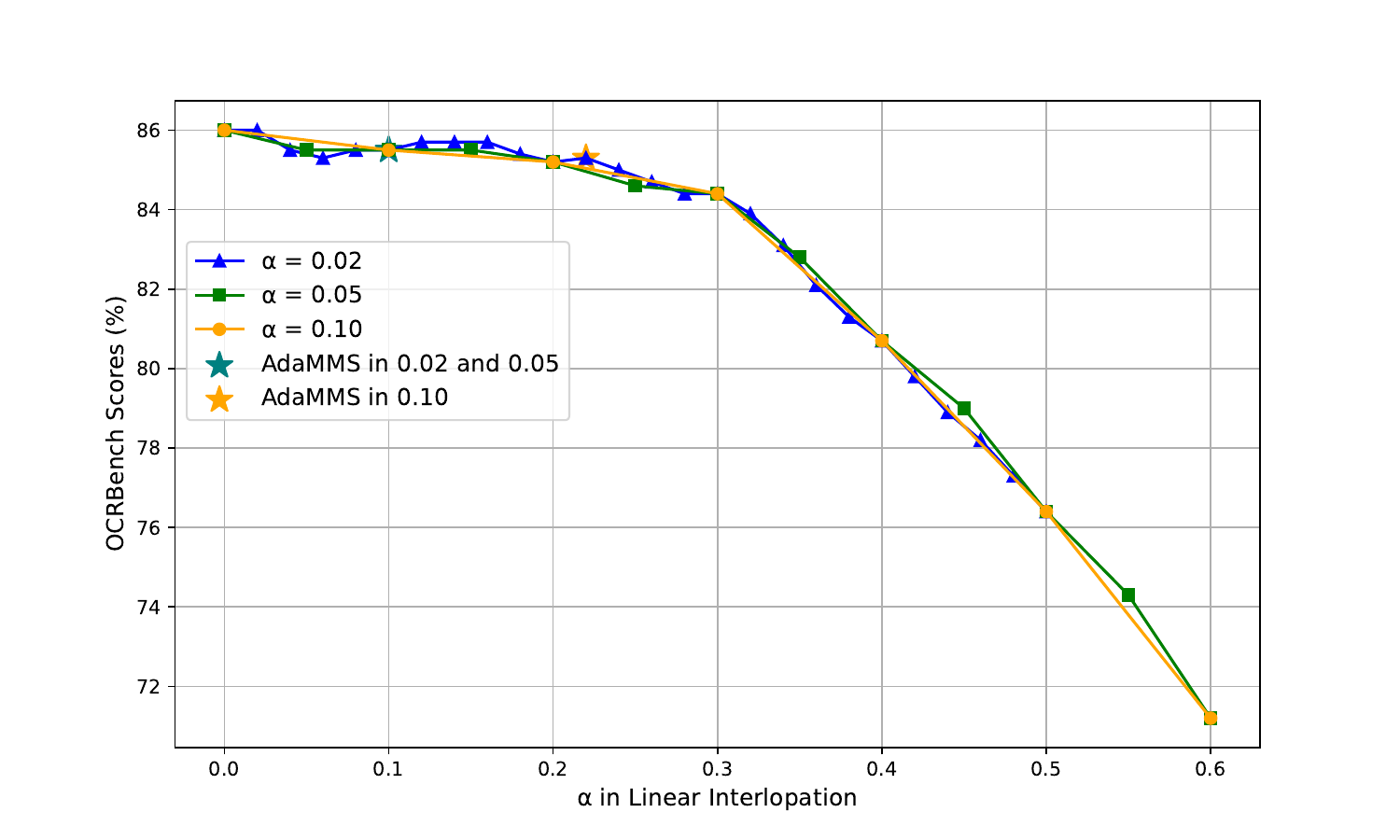}
    \end{subfigure}
    \caption{Results on linear interpolation at different granularities of $\alpha$ when merging LLaVA-OneVison-7B into Qwen2-VL-7B-7B. (Left: MME, Right: OCRBench)}
    \label{fig:step_size}
\end{figure*}

\begin{figure*}[h]
    \centering
    \includegraphics[width=\linewidth, bb=0 0 1007 576]{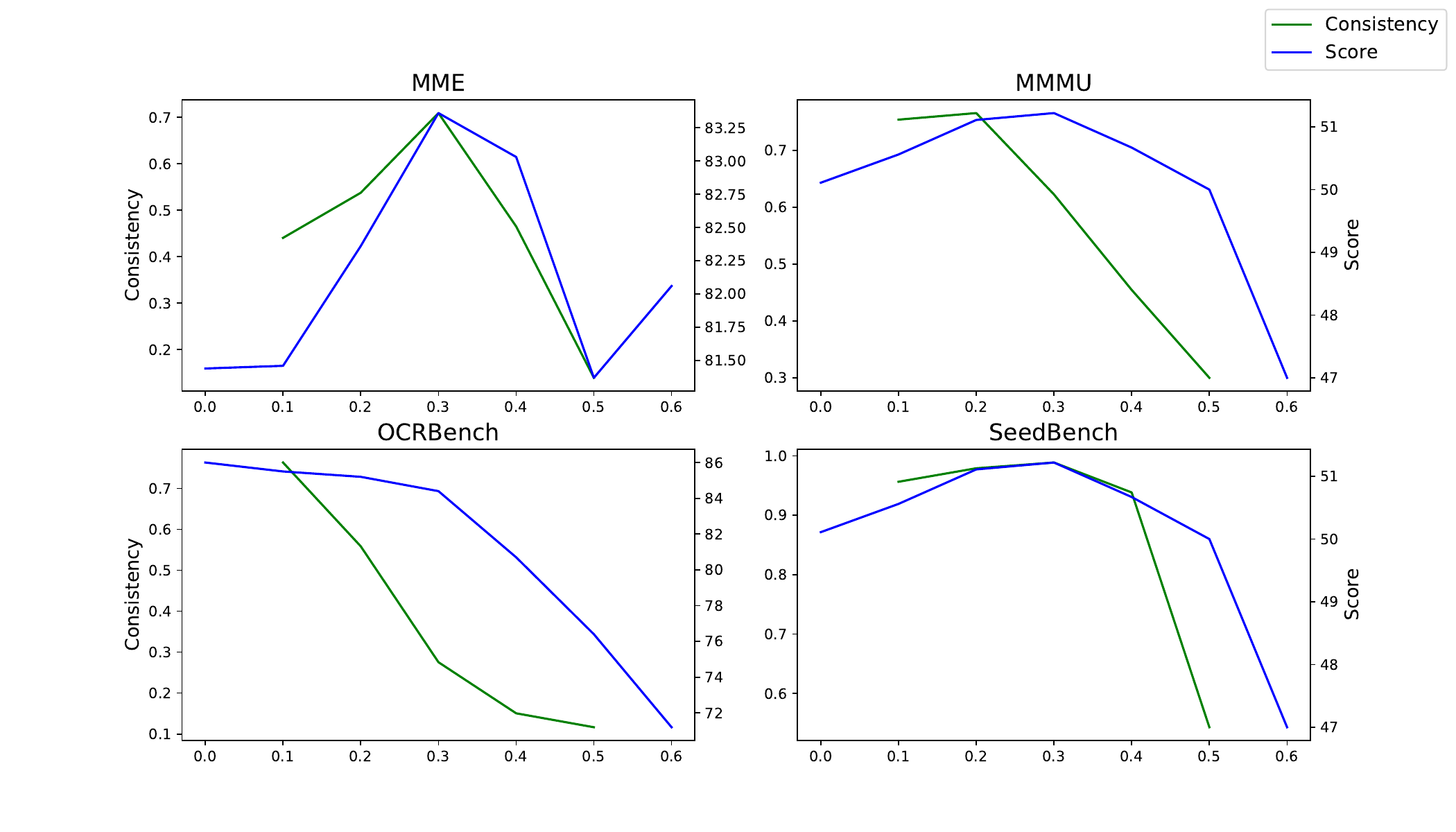}
    \caption{Generation consistency and model performance (score) for MME, MMMU, OCRBench and SeedBench when merging LLaVA-OneVision-7B into Qwen2-VL-7B. Generation consistency is calculated as the reciprocal of the sum of different responses from models with adjacent $\alpha$ candidates. The horizontal axis is the $\alpha$ of the linear interpolation.}
    \label{fig:consistency_acc4}
\end{figure*}

Figure~\ref{fig:step_size} presents the result of \ours at different granularities of $\alpha$. The point in stars indicates the best $\alpha$ by our unsupervised parameter selection method. The result shows that these granualities in $\{0.02, 0.05, 0.10\}$ behave similarly in terms of the final performance, indicating the robustness of \ours. Therefore, in practice we choose a larger $\alpha$ so that we have fewer $\alpha$ candidates, which reduces the computation cost.

\begin{table*}[t]
    \centering
    
     \resizebox{\textwidth}{!}{
    \begin{tabular}{lllllllllll}
    \hline
        Model & $\mathrm{MMMU_{val}}$ & $\mathrm{MME_{sum}}$ & $\mathrm{SeedBench_{all}}$ & $\mathrm{OCRBench}$ & $\mathrm{TextVQA_{val}}$ & $\mathrm{OKVQA}$ & $\mathrm{GQA}$ & $\mathrm{VizWiz_{val}}$ & $\mathrm{Sum}$ & $\mathrm{Diff}$  \\ \hline
        AdaMMS & 34.90  & 69.09  & 64.12  & 55.70  & 76.90  & 61.11  & 60.12  & 37.27  & 459.21  & +31.23 \\ 
        Ties-Merging  & 34.00  & 57.29  & 38.97  & 55.00  & 59.73  & 40.31  & 51.97  & 24.36  & 361.63  & -66.35 \\ 
        Ties-Merging (supervised with 100 eval. samples) & 37.20  & 57.29  & 63.12  & 55.90  & 76.50  & 61.45  & 55.81  & 37.98  & 445.25  & +17.27 \\ 
        Ties-Merging (supervised with all eval. data) & 37.20  & 63.96  & 65.43  & 55.90  & 76.55  & 61.45  & 57.99  & 38.21  & 456.69  & +28.71 \\ 
        \hline
        
\end{tabular}
    }
    
    
    \captionof{table}{ \ours and Ties-Merging with \textit{supervised} hyper-parameter selection via validation set.}

    
    \label{tab:supervised}

    

\end{table*}

\begin{table*}[!ht]
    \centering
       \resizebox{\textwidth}{!}{%
    
    \begin{tabular}{lrrrrrrrrrr}
    \toprule
        Model & $\mathrm{MMMU_{val}}$ &  $\mathrm{MME_{sum}}$ &  $\mathrm{SeedBench_{all}}$ & $\mathrm{OCRBench}$  &  $\mathrm{TextVQA_{val}}$  & $\mathrm{OKVQA}$ & $\mathrm{GQA}$  &  $\mathrm{VizWiz_{val}}$ & $\mathrm{SUM}$ & $\mathrm{Top2}$ \\ 
        \midrule     
       
         Task Arithmetic & \underline{13.21} & \textbf{21.53} & \textbf{14.54} & -1.80 & -3.74 & \textbf{13.88} & \underline{-2.95} & \underline{-7.29} & \underline{47.45} & 7 \\ 
        Ties-Merging & -3.32 & -24.94 & -27.34 & 1.20 & -31.59 & -23.23 & -29.70 & -29.20 & -168.05 & 0 \\ 
        DARE-Linear & 8.15 & -2.35 & 10.58 & -12.3 & -19.23 & 5.41 & -12.76 & -21.09 & -43.53 & 0 \\ 
        DARE-Ties & -14.83 & -60.56 & -6.96 & -47.50 & -47.12 & -31.08 & -26.32 & -32.45 & -266.76 & 0 \\ 
        MetaGPT & 1.44 & -2.93 & -4.02 & \textbf{15.30} & \underline{0.37} & -6.75 & -23.69 & -16.73 & -36.94 & 2 \\ 
        AdaMMS & \textbf{17.68} & \underline{17.48} & \underline{12.02} & \underline{13.60} & \textbf{18.43} & \underline{13.40} & \textbf{1.40} & \textbf{-2.14} &\textbf{ 91.92} & 9 \\  \bottomrule

    \end{tabular}
    }
    \caption{Results of the performance gain sum among six model pairs reported in our paper, as described in Appendix~\ref{appendix:additional}. The performance gain for each task is computed as the difference between the performance of our method (or baselines) and the average performance of the two original models, with positive values indicating an improvement.}
     \label{tab:avg6}
\end{table*}

\begin{table*}[!ht]
    \centering
    \resizebox{\textwidth}{!}{%
    \begin{tabular}{lclllllllllc}
        \toprule             

        Model & $\mathrm{Unsupervised}$ & $\mathrm{MMMU_{val}}$ &  $\mathrm{MME_{sum}}$ &  $\mathrm{SeedBench_{all}}$ & $\mathrm{OCRBench}$  &  $\mathrm{TextVQA_{val}}$  & $\mathrm{OKVQA}$ & $\mathrm{GQA}$  &  $\mathrm{VizWiz_{val}}$ & $\mathrm{SUM}$  & $\mathrm{Top2}$  \\ 
        \midrule

\rowcolor{gray!20}
\multicolumn{12}{c}{\textbf{Original Models}} \\
\midrule
 LLaVA\footnotesize(base) & ~ & 35.10  & 66.68  & 60.52  & 31.30  & 46.04  & 53.42  & 61.94  & 54.29  & 409.29 \\
        mPLUG-Owl2 & ~ & 34.90  & 62.80  & 59.41  & 34.10  & 55.13  & 60.98  & 56.11  & 32.07   & 395.50  \\ 
        
     \hline
       
\rowcolor{gray!20}
\multicolumn{12}{c}{\textbf{Baselines}} \\
\hline
  Task Arithmetic & $\times$ & \underline{36.00 (+1.00)} & 67.00 (+2.26) & 61.45 (+1.48) & 30.40 (-2.30) & 45.75 (-4.84) & \underline{56.79 (-0.41)} & \underline{59.68 (+0.66)} & \underline{56.49 (+13.31)} & \underline{413.56 (+11.17)} & 5 \\ 
        Ties-Merging &  $\times$ & 33.60 (-1.40) & 62.14 (-2.60) & 60.32 (+0.35) & 30.10 (-2.60) & 42.85 (-7.73) & 52.46 (-4.74) & 58.37 (-0.66) & 51.30 (+8.12) & 391.14 (-11.25) & 0 \\ 
        
        DARE-Linear &  $\times$ & 36.00 (+1.00) & 67.00 (+2.26) & 61.41 (+1.44) & 30.70 (-2.00) & \underline{45.84 (-4.74)} & \textbf{57.06 (-0.14)} & 59.56 (+0.54) & 55.90 (+12.72) & 413.47 (+11.08) & 2\\ 
        DARE-Ties &  $\times$ & 31.70 (-3.30) & 59.81 (-4.93) & 60.06 (+0.09) & 29.50 (-3.20) & 41.90 (-8.69) & 46.00 (-11.20) & 57.51 (-1.52) & 53.27 (+10.09) & 379.75 (-22.64) & 0 \\ 
        MetaGPT &  $\checkmark$ & 35.30 (+0.30) & \textbf{67.62 (+2.88)} & \underline{61.46 (+1.49)} & \underline{30.60 (-2.10)} & 45.80 (-4.79) & 56.54 (-0.66) & 59.41 (+0.38) & \textbf{56.66 (+13.48)} & 413.39 (+11.00) & 4 \\[0.5ex] 
        
       \hline
       
\rowcolor{gray!20}
\multicolumn{12}{c}{\textbf{Our Method}} \\
\hline 
                      
        AdaMMS &$\checkmark$& \textbf{38.30 (+3.30)} & \underline{67.01 (+2.27)} & \textbf{61.82 (+1.85)} & \textbf{31.00 (-1.70)} & \textbf{46.49 (-4.09)} & 55.60 (-1.60) &\textbf{ 61.81 (+2.79)} & 54.64 (+11.46) & \textbf{416.67 (+14.28)} & 7 \\         \bottomrule
    \end{tabular}%
        }
    \caption{Results on merging mPLUG-Owl2-7B into LLaVA-v1.5-7B.}
    \label{tab:mplug2llava}
\end{table*}

\begin{table*}[!ht]
    \centering
    \resizebox{\textwidth}{!}{%
    \begin{tabular}{lclllllllllc}
        \toprule             

        Model & $\mathrm{Unsupervised}$ & $\mathrm{MMMU_{val}}$ &  $\mathrm{MME_{sum}}$ &  $\mathrm{SeedBench_{all}}$ & $\mathrm{OCRBench}$  &  $\mathrm{TextVQA_{val}}$  & $\mathrm{OKVQA}$ & $\mathrm{GQA}$  &  $\mathrm{VizWiz_{val}}$ & $\mathrm{SUM}$  & $\mathrm{Top2}$  \\ 
        \hline

\rowcolor{gray!20}
\multicolumn{12}{c}{\textbf{Original Models}} \\
\hline
        mPLUG-Owl2\footnotesize(base) & ~ & 34.90  & 62.80  & 59.41  & 34.10  & 55.13  & 60.98  & 56.11  & 32.07   & 395.50  \\ 
        LLaVA & ~ & 35.10  & 66.68  & 60.52  & 31.30  & 46.04  & 53.42  & 61.94  & 54.29  & 409.29 \\
     \hline
       
\rowcolor{gray!20}
\multicolumn{12}{c}{\textbf{Baselines}} \\
\hline
 Task Arithmetic & $\times$ & \underline{36.90\footnotesize(+1.90)} & 63.17\footnotesize(-1.57) & \textbf{60.44\footnotesize(+0.47)} & 33.00\footnotesize(+0.30) & 55.40\footnotesize(+4.81) & \textbf{63.87\footnotesize(+6.67)} & 56.97\footnotesize(-2.06) & \textbf{33.70\footnotesize(-9.48)} & 403.45\footnotesize(+1.05) & 4\\ 
 
        Ties-Merging & $\times$ & \underline{36.90\footnotesize(+1.90)} & 64.20\footnotesize(-0.54) & 60.13\footnotesize(+0.16) & \textbf{34.40\footnotesize(+1.70)} & 54.50\footnotesize(+3.91) & 62.92\footnotesize(+5.72) & \textbf{57.55\footnotesize(-1.48)} & 33.18\footnotesize(-10.00) & \textbf{403.78\footnotesize(+1.38)} &4 \\
        
        DARE-Linear & $\times$  &36.20\footnotesize(+1.20) & 62.99\footnotesize(-1.75) & \underline{60.41\footnotesize(+0.44)} & 32.60\footnotesize(-0.10) & 55.15\footnotesize(+4.56) & \underline{63.47\footnotesize(+6.27)} & 56.73\footnotesize(-2.30) & 33.35\footnotesize(-9.83) & 400.90\footnotesize(-1.50) & 2 \\ 
        
        DARE-Ties & $\times$ & 35.30\footnotesize(+0.30) & 60.37\footnotesize(-4.37) & 58.36\footnotesize(-1.61) & 32.00\footnotesize(-0.70) & 51.65\footnotesize(+1.06) & 58.08\footnotesize(+0.88) & 55.57\footnotesize(-3.46) & 31.03\footnotesize(-12.15) & 382.36\footnotesize(-20.04) &0 \\ 
        
        MetaGPT & $\checkmark$ & 36.00\footnotesize(+1.00) & \underline{64.24\footnotesize(-0.50)} & 60.23\footnotesize(+0.26) & \underline{33.90\footnotesize(+1.20)} & \underline{55.83\footnotesize(+5.24)} & 62.88\footnotesize(+5.68) & 56.53\footnotesize(-2.50) & 33.35\footnotesize(-9.83) & 402.96\footnotesize(+0.56) & 3
          \\[0.5ex] 
        
       \hline
       
\rowcolor{gray!20}
\multicolumn{12}{c}{\textbf{Our Method}} \\
\hline 
                      
       AdaMMS &$\checkmark$& \textbf{37.60\footnotesize(+2.60)} & \textbf{64.61\footnotesize(-0.13)} & 60.02\footnotesize(+0.05) & 32.20\footnotesize(-0.50) & \textbf{55.84\footnotesize(+5.25)} & 63.13\footnotesize(+5.93) & \underline{56.98\footnotesize(-2.05)} & \underline{33.39\footnotesize(-9.79)} & \underline{403.77\footnotesize(+1.37)} & 6\\
        \bottomrule
    \end{tabular}%
        }
    \caption{Results on merging LLaVA-v1.5-7B into mPLUG-Owl2-7B.}
    \label{tab:llava2mplug}
\end{table*}

\begin{table*}[!ht]
    \centering
    \resizebox{\textwidth}{!}{%
    \begin{tabular}{lclllllllllc}
        \toprule             

        Model & $\mathrm{Unsupervised}$ & $\mathrm{MMMU_{val}}$ &  $\mathrm{MME_{sum}}$ &  $\mathrm{SeedBench_{all}}$ & $\mathrm{OCRBench}$  &  $\mathrm{TextVQA_{val}}$  & $\mathrm{OKVQA}$ & $\mathrm{GQA}$  &  $\mathrm{VizWiz_{val}}$ & $\mathrm{SUM}$  & $\mathrm{Top2}$  \\ 
        \hline

\rowcolor{gray!20}
\multicolumn{12}{c}{\textbf{Original Models}} \\
\hline
 mPLUG-Owl2\footnotesize(base) & ~ & 34.90  & 62.80  & 59.41  & 34.10  & 55.13  & 60.98  & 56.11  & 32.07  & 395.50  \\ 
     CogVLM & ~  & 34.80  & 59.23  & 61.22  & \textit{56.50}  & \textit{77.57}  & 60.82  & \textit{59.43}  & \textit{37.09}  & \textit{446.66} \\

     \hline
       
\rowcolor{gray!20}
\multicolumn{12}{c}{\textbf{Baselines}} \\
\hline
Task Arithmetic &$\times$ & \underline{38.80\footnotesize(+3.95)} & \textbf{64.65\footnotesize(+3.63)} & \textbf{60.85\footnotesize(+0.53)} & \textbf{31.50\footnotesize(-13.80)} & \textbf{56.99\footnotesize(-9.36)} & 60.93\footnotesize(+0.03) & \underline{54.44\footnotesize(-3.33)} & \textbf{32.76\footnotesize(-1.82)} & \textbf{400.92\footnotesize(-20.16)} & 8 \\

        Ties-Merging &$\times$ & 27.9\footnotesize(-6.95) & 48.96\footnotesize(-12.06) & 52.32\footnotesize(-8.00) & 24.30\footnotesize(-21.00) & 42.10\footnotesize(-24.25) & 54.15\footnotesize(-6.75) & 43.02\footnotesize(-14.75) & 27.56\footnotesize(-7.02) & 320.31\footnotesize(-100.77) &0 \\ 
        
        DARE-Linear & $\times$ &37.60\footnotesize(+2.75) & 62.44\footnotesize(+1.42) & 59.81\footnotesize(-0.51) & 30.90\footnotesize(-14.40) & 56.41\footnotesize(-9.94) & \underline{61.07\footnotesize(+0.17)} & 54.11\footnotesize(-3.66) & 32.42\footnotesize(-2.16) & 394.76\footnotesize(-26.32) & 1\\ 
        
        DARE-Ties & $\times$ &32.00\footnotesize(-2.85) & 57.90\footnotesize(-3.12) & 57.62\footnotesize(-2.70) & 24.10\footnotesize(-21.20) & 43.84\footnotesize(-22.51) & 51.56\footnotesize(-9.34) & 52.04\footnotesize(-5.73) & 25.67\footnotesize(-8.91) & 344.73\footnotesize(-76.35) & 0\\
        
        MetaGPT & $\checkmark$&31.30\footnotesize(-3.55) & 56.81\footnotesize(-4.21) & 50.81\footnotesize(-9.51) & 29.30\footnotesize(-16.00) & 37.96\footnotesize(-28.39) & 43.02\footnotesize(-17.88) & 34.12\footnotesize(-23.65) & 15.84\footnotesize(-18.74) & 299.16\footnotesize(-121.92)  &0
          \\[0.5ex] 
               \hline       
\rowcolor{gray!20}
\multicolumn{12}{c}{\textbf{Our Method}} \\
\hline                    
       AdaMMS &$\checkmark$& \textbf{39.10\footnotesize(+4.25)} & \textbf{64.65\footnotesize(+3.63)} & \underline{60.16\footnotesize(-0.16)} & \underline{30.60\footnotesize(-14.70)} & \underline{55.88\footnotesize(-10.47)} & \textbf{62.11\footnotesize(+1.21)} & \textbf{55.61\footnotesize(-2.16)} & \underline{32.69\footnotesize(-1.89)} & \underline{400.80\footnotesize(-20.28)} & 9\\
        \bottomrule
    \end{tabular}%
        }
    \caption{Results on merging CogVLM-7B into mPLUG-Owl2-7B.}
    \label{tab:cog2mplug}
\end{table*}

\begin{table*}[!ht]
    \centering
    \resizebox{\textwidth}{!}{%
    \begin{tabular}{lclllllllllc}
        \toprule             

        Model & $\mathrm{Unsupervised}$ & $\mathrm{MMMU_{val}}$ &  $\mathrm{MME_{sum}}$ &  $\mathrm{SeedBench_{all}}$ & $\mathrm{OCRBench}$  &  $\mathrm{TextVQA_{val}}$  & $\mathrm{OKVQA}$ & $\mathrm{GQA}$  &  $\mathrm{VizWiz_{val}}$ & $\mathrm{SUM}$  & $\mathrm{Top2}$  \\ 
        \hline

\rowcolor{gray!20}
\multicolumn{12}{c}{\textbf{Original Models}} \\
\hline
          CogVLM\footnotesize(base) & ~ &  34.80  &  59.23  &  61.22  &  56.50  &  77.57  &  60.82  &  59.43  &  37.09  &  446.66  \\ 
        mPLUG-OWI2 & ~ &  34.90  &  62.80  &  59.41  &  34.10  &  55.13  &  60.98  &  56.11  &  32.07  &  395.50 \\
     \hline
       
\rowcolor{gray!20}
\multicolumn{12}{c}{\textbf{Baselines}} \\
\hline
 Task Arithmetic &$\times$ & \underline{38.30\footnotesize(+3.45)} & \textbf{72.11\footnotesize(+11.09)} & \textbf{67.24\footnotesize(+6.92)} & 51.90\footnotesize(+6.60) & 70.68\footnotesize(+4.33) & \textbf{63.59\footnotesize(+2.69)} & \textbf{59.98\footnotesize(+2.21)} & \underline{37.16\footnotesize(+2.58)} &\textbf{460.96\footnotesize(+39.88)} &7 \\ 
 
        Ties-Merging &$\times$ & 34.60\footnotesize(-0.25) & 53.54\footnotesize(-7.48) & 61.73\footnotesize(+1.41) & 50.70\footnotesize(+5.40) & 66.65\footnotesize(+0.30) & 58.19\footnotesize(-2.71) & 52.66\footnotesize(-5.11) & 33.92\footnotesize(-0.66) & 411.99\footnotesize(-9.09) &0\\ 
        
        DARE-Linear & $\times$ &\textbf{39.20\footnotesize(+4.35) }& \underline{68.80\footnotesize(+7.78) }& 66.66\footnotesize(+6.34) & 50.90\footnotesize(+5.60) & 70.35\footnotesize(+4.00) & \underline{63.26\footnotesize(+2.36)} & 58.80\footnotesize(+1.03) & 36.80\footnotesize(+2.22) & 454.77\footnotesize(+33.69)&3 \\ 
        
        DARE-Ties & $\times$ &29.00\footnotesize(-5.85) & 53.89\footnotesize(-7.12) & 61.61\footnotesize(+1.30) & 42.90\footnotesize(-2.40) & 63.46\footnotesize(-2.89) & 54.34\footnotesize(-6.56) & 55.54\footnotesize(-2.23) & 33.96\footnotesize(-0.62) & 394.70\footnotesize(-26.38)&0 \\ 
        
        MetaGPT & $\checkmark$& 34.90\footnotesize(+0.05) & 61.54\footnotesize(+0.52) & 62.93\footnotesize(+2.62) & \textbf{57.30\footnotesize(+12.00)} & \textbf{77.18\footnotesize(+10.83)} & 61.55\footnotesize(+0.65) & 59.93\footnotesize(+2.16) & 37.15\footnotesize(+2.57) & 452.48\footnotesize(+31.40) &2
          \\[0.5ex] 
        
       \hline
       
\rowcolor{gray!20}
\multicolumn{12}{c}{\textbf{Our Method}} \\
\hline 
                      
      AdaMMS &$\checkmark$& 38.10\footnotesize(+3.25) & 62.48\footnotesize(+1.46) & \underline{66.79\footnotesize(+6.48)} & \underline{56.30\footnotesize(+11.00)} & \underline{76.89\footnotesize(+10.54)} & 61.71\footnotesize(+0.81) & \underline{59.96\footnotesize(+2.19)} & \textbf{37.33\footnotesize(+2.75)} & \underline{459.56\footnotesize(+38.48)} &6  \\
        \bottomrule
    \end{tabular}%
        }
    \caption{Results on merging mPLUG-Owl2-7B into CogVLM-7B.}
    \label{tab:mplug2cog}
\end{table*}

\begin{table*}[!ht]
    \centering
    \small
    \resizebox{\textwidth}{!}{%
    \begin{tabular}{cccccccccc}
    \toprule
          Model & $\mathrm{MMMU_{val}}$ &  $\mathrm{MME_{sum}}$ &  $\mathrm{SeedBench_{all}}$ & $\mathrm{OCRBench}$  &  $\mathrm{TextVQA_{val}}$  & $\mathrm{OKVQA}$ & $\mathrm{GQA}$  &  $\mathrm{VizWiz_{val}}$ & $\mathrm{SUM}$    \\ 
        
          \hline
\rowcolor{gray!20}
\multicolumn{10}{c}{\textbf{Original Models}} \\
\hline
         Qwen2-VL\footnotesize(base) & 50.11  & 81.44  & 75.85  & 86.00  & 84.12  & 51.43  & 61.80  & 68.32  & 559.07  \\ 
         LLaVA-OneVision  & 43.44  & 77.04  & 75.44  & 69.60  & 78.47  & 49.57  & 59.84  & 60.97  & 514.37  \\ 
        AVG & 46.78  & 79.24  & 75.65  & 77.80  & 81.30  & 50.50  & 60.82  & 64.65  & 536.72  \\ 
                  \hline
\rowcolor{gray!20}
\multicolumn{10}{c}{\textbf{Linear Interpolation}} \\
\hline
        $\alpha$-0.1 & 50.56  & 81.46  & 76.20  & \textbf{85.50}  & \textbf{83.41}  & 53.56  & 62.02  & \textbf{68.40}  & \textbf{561.11}  \\ 
        $\alpha$-0.2 & 51.11  & 82.36  & 76.23  & 85.20  & 81.74  & \textbf{54.76}  & \textbf{62.05}  & 67.12  & 560.57  \\ 
        $\alpha$-0.3 & \textbf{51.22}  & \textbf{83.36}  & \textbf{76.34}  & 84.40  & 78.43  & 52.03  & 61.44  & 63.91  & 551.13  \\ 
        $\alpha$-0.4 & 50.67  & 83.03  & 76.06  & 80.70  & 71.66  & 49.83  & 60.09  & 58.43  & 530.47  \\ 
        $\alpha$-0.5 & 50.00  & 81.37  & 75.63  & 76.40  & 59.13  & 44.96  & 55.53  & 52.60  & 495.62  \\ 
        $\alpha$-0.6 & 47.00  & 82.06  & 74.76  & 71.30  & 39.37  & 40.31  & 54.11  & 46.39  & 455.30  \\ 
       
          \hline
\rowcolor{gray!20}
\multicolumn{10}{c}{\textbf{Our Method}} \\
\hline
        AdaMMS & 51.11  & 83.36  & 76.20  & 85.50  & 83.41  & 53.56  & 62.02  & 68.40  & 563.56  \\ \midrule
        Selected $\alpha$ & 0.2 & 0.3 & 0.1 & 0.1 & 0.1 & 0.1 & 0.1 & 0.1 & - \\ 
        Distance with the best $\alpha$ & 0.1 & 0 & 0.2 & 0 & 0 & 0.1 & 0.1 & 0 & - \\ \bottomrule
    \end{tabular}
    }
    \caption{Intermediate results on different $\alpha$ candidates in the linear interpolation of AdaMMS, and the $\alpha$ selected by our unsupervised hyper-parameter selection method on merging LLaVA-OneVision-7B into Qwen2-VL-7B. AVG indicates the average performance of the two original models.}
    \label{tab:inter_alpha}
\end{table*}

\end{document}